\pdfoutput=1

\documentclass[11pt,dvipsnames,table,xcdraw]{article}

\usepackage[]{emnlp2021}

\usepackage{times}
\usepackage{latexsym}

\usepackage{amsmath}

\usepackage[T1]{fontenc}

\usepackage[utf8]{inputenc}

\usepackage{microtype}

\usepackage{tikz}
\usepackage{caption}
\usetikzlibrary{positioning,shapes,shadows,arrows}
\usepackage{graphicx}
\usepackage{multirow}
\usepackage{subcaption}
\usepackage{multicol}
\usepackage{array}
\newcolumntype{L}[1]{>{\raggedright\let\newline\\\arraybackslash\hspace{0pt}}m{#1}}
\newcolumntype{C}[1]{>{\centering\let\newline\\\arraybackslash\hspace{0pt}}m{#1}}
\newcolumntype{R}[1]{>{\raggedleft\let\newline\\\arraybackslash\hspace{0pt}}m{#1}}
\usepackage{lscape}
\usepackage{hhline}
\usepackage{bbm}
\usepackage{array}
\usepackage{longtable}
\usepackage{colortbl}

\usepackage{enumitem} %

\title{Perturbation CheckLists for Evaluating NLG Evaluation Metrics}

\author{
Ananya B. Sai$^{1,2}$ ~~ Tanay Dixit$^{1}$ ~~ Dev Yashpal Sheth$^{1,2}$ \\ %
\textbf{Sreyas Mohan}$^{3}$ ~~ \textbf{Mitesh M. Khapra}$^{1,2,4}$ \\
$^{1}$ Indian Institute of Technology, Madras \\
$^{2}$ Robert Bosch Centre for Data Science and Artificial Intelligence, IIT Madras \\
$^{3}$ Center for Data Science, New York University ~~~~
$^{4}$ AI4Bharat \\
\texttt{\{ananya,devsheth,miteshk\}@cse.iitm.ac.in,}
\\
\texttt{dixittanay@gmail.com , sm7582@nyu.edu}
}

\definecolor{color}{RGB}{191, 239, 255}
\definecolor{lightred}{RGB}{255, 204, 204}

\begin{document}
\maketitle
\begin{abstract}
Natural Language Generation (NLG) evaluation is a multifaceted task requiring assessment of multiple desirable criteria, \textit{e.g.}, fluency, coherency, coverage, relevance, adequacy, overall quality, \textit{etc.} Across existing datasets for 6 NLG tasks, we observe that the human evaluation scores on these multiple criteria are often not correlated. For example, there is a very low correlation between human scores on \textit{fluency} and \textit{data coverage} for the task of structured data to text generation. This suggests that the current recipe of proposing new automatic evaluation metrics for NLG by showing that they correlate well with scores assigned by humans for a \textit{single} criteria (overall quality) alone is inadequate. Indeed, our extensive study involving 25 automatic evaluation metrics across 6 different tasks and 18 different evaluation criteria shows that there is no single metric which correlates well with human scores on \textit{all} desirable criteria, for most NLG tasks. 
Given this situation, we propose \textit{CheckLists} for better design and evaluation of automatic metrics. We design templates which target a specific criteria (e.g., coverage) and perturb the output such that the quality gets affected only along this specific criteria (e.g., the coverage drops). We show that existing evaluation metrics are not robust against even such simple perturbations and disagree with scores assigned by humans to the perturbed output. The proposed templates thus allow for a fine-grained assessment of automatic evaluation metrics exposing their limitations and will facilitate better design, analysis and evaluation of such metrics.\footnote{Our templates and code are available at \href{https://iitmnlp.github.io/EvalEval/}{https://iitmnlp.github.io/EvalEval/}} %
\end{abstract}

\section{Introduction}
As the number of tasks and benchmarks for NLG have increased \cite{DBLP:journals/corr/abs-2102-01672_gem_benchmark}, the challenges in evaluating NLG systems have also continued to grow \cite{DBLP:conf/emnlp/LiuLSNCP16HowNot,DBLP:conf/emnlp/NemaK18_qbleu,DBLP:conf/aaai/SaiGKS19Reeval}. %
One reliable way of evaluating NLG systems is to collect human judgements. However, this is a time consuming and expensive process \citep{DBLP:journals/corr/abs-2104-14478,DBLP:journals/corr/abs-1905-04071,howcroft-etal-2020-twenty}.
Hence, automatic evaluation metrics such as BLEU \cite{bleu} which are quicker to compute have become popular, despite being less reliable \cite{callison-burch-etal-2006-evaluating_bleu_in_mt, DBLP:journals/coling/Reiter18_validity_of_bleu}.

The survey by \citet{DBLP:journals/corr/abs-2008-12009_NLG_evaluation_metrics_survey} shows that more than 35 automatic evaluation metrics have been proposed for NLG since 2014, 
however, there is no careful evaluation of the ability of such metrics to assess the quality of the output of an NLG system on multiple desired criteria. %
For example, consider the task of dialog evaluation, where humans are asked to score the output on multiple criteria such as \textit{fluency, adequacy, coherence, informativeness, engagingness, consistency, etc}. Contrast this with automatic evaluation metrics such as BLEU, BLEURT \cite{DBLP:conf/acl/SellamDP20_bleurt}, DEB \cite{DBLP:journals/tacl/SaiMAK20_DDpp}, ADEM \cite{DBLP:conf/acl/LoweNSABP17ADEM}, etc., which assign a \textit{single} score to the output. What does this score indicate? More specifically, does a low DEB score indicate that the output is not fluent or does it indicate that the output is fluent but not coherent or neither fluent nor coherent? Hence, a single overall score assigned by automatic evaluation metrics %
 is not very informative in deciding which aspects of model improvement should one focus on.

The question then is why do current automatic evaluation metrics produce only a single overall score?
This is simply because of a conscious choice made while designing automatic evaluation metrics. In particular, current works only focus on evaluating whether the scores assigned by the proposed metric correlate well with the \textit{overall quality} scores assigned by humans as opposed to \textit{all} relevant criteria. %
In this work, we make a case for shifting the focus to \textit{all} relevant criteria while evaluating such metrics. To this end, we first do a systematic study involving 6 NLG tasks, 18 different human evaluation criteria (\textit{fluency, coverage,coherence, consistency, etc}) and 25 automatic evaluation metrics. We take existing English datasets containing human judgements for various tasks and criteria and make two important observations. First, for a given task, human scores for different criteria often have a low correlation, thereby suggesting that these criteria cannot be clubbed together and evaluated using a single score assigned by an automatic evaluation metric. Second, none of the automatic evaluation metrics have a high correlation with human scores for any of the desired criteria for a given task. 

The above results highlight a lacunae in the evaluation of automatic evaluation metrics wherein their ability to assess the output on multiple criteria is not evaluated. In this work, we propose a flexible framework which allows a systematic evaluation of the capabilities of an automatic evaluation metric. 
In particular, we propose CheckList-style templates \cite{DBLP:conf/acl/RibeiroWGS20_checklist} which evaluate the robustness of the metrics to certain perturbations targeting specific criteria.  We illustrate this idea with an example in Table~\ref{tab:example_intro_perturb}. In row 2 of Table~\ref{tab:example_intro_perturb} a gold standard output is perturbed by changing named entities, thereby affecting its \textit{factual correctness} which is important for data-to-text generation. If an automatic evaluation metric indeed evaluates \textit{factual correctness} then its score should drop when presented with such a perturbed output. %
\begin{table}[!t]
\resizebox{\columnwidth}{!}{%
\begin{tabular}{|l l|}
\hline
Original sentence:       &                {\color[HTML]{009901} Cameron} is the director of Titanic        \\
\hline
Change names: & {\color[HTML]{CB0000} Kate} is the director of Titanic (incorrect) \\
\hline
\end{tabular}%
}
\caption{Example of a perturbed output.}
\label{tab:example_intro_perturb}
\end{table}

For the 6 NLG tasks mentioned earlier, we create 34 such perturbation templates covering 18 different evaluation criteria.We then instantiate these templates to create large-scale test cases. For every perturbation, we also collect human judgements to understand how much would a human change his/her score when shown such a perturbed output. %
We find that for several perturbations, the scores assigned by automatic evaluation metrics do not agree with the scores assigned by humans, thereby indicating that current automatic evaluation metrics are not robust to such perturbations (i.e., they do not really evaluate the desired criteria). %
Overall, we believe that the proposed templates provide a better framework for a more fine-grained evaluation of automatic evaluation metrics which goes much beyond computing correlations with human scores.

\begin{table*}[!ht]
\resizebox{1\textwidth}{!}{
\small
\begin{tabular}{L{3.2cm}|L{13.9cm}}
\hline\hline
\textbf{Task}                                       & \textbf{Criteria}                                                                                                                                                \\ \hline\hline
Machine Translation                        & \textbf{Adequacy:} The generated translation should adequately represent all the information present in the reference.                                  \\ \hline
\multirow{2}{*}{Question Generation}       & \textbf{Relevance}: Is the question related to the source material they are based upon.                                                                 \\ %
                                           & \textbf{Answerability}: Is the generated question answerable given the context.                                                                         \\ \hline
& \textbf{Informativeness:} The summary should convey the key points of the text.                                                                         \\ %
                                           & \textbf{Non-redundancy:} The summary should not repeat any points, and ideally have maximal information coverage within the limited text length.        \\ %
 Abstractive Summarization                
        & \textbf{Referential clarity:} Any intra-sentence or cross-sentence references in the summary should be unambiguous and within the scope of the summary. \\ %
            & \textbf{Focus:} The summary needs to have a focus and all the sentences need to contain information related to this focal point.                        \\ %
                                           & \textbf{Structure and Coherence:} The summary should be a well-organized and coherent body of information                                               \\ \hline
\multirow{7}{*}{Dialogue Generation}      & \textbf{Making sense:} Does the bot say things that don't make sense?                                                                                   \\ %
                                           & \textbf{Engagingness:} Is the dialogue agent enjoyable to talk to?                                                                                      \\ %
                                           & \textbf{Interestingness:} Did you find the bot interesting to talk to?                                                                                  \\ %
                                           & \textbf{Inquisitivenes:} Does the bot ask a good amount of questions?                                                                                   \\ %
                                           & \textbf{Listening:} Does the bot pay attention to what you say?                                                                                         \\ %
                                           & \textbf{Avoiding Repetition:} Does the bot repeat itself? (either within or across utterances)                                                          \\ %
                                           & \textbf{Humanness:} Is the conversation with a person or a bot?                                                                                         \\ \hline
\multirow{2}{*}{Image Captioning}       & \textbf{Relevance}: The caption should be specific and related to the image.                                                            \\ %
                                           & \textbf{Thoroughness}: The caption should adequately describe the image.                                                                      \\ \hline
& \textbf{Data  Coverage}:  Does  the  text  include  descriptions of all predicates presented in the data?                                                     \\ %
                                           & \textbf{Relevance}: Does the text describe only such predicates %
                                           which are found in the data?                                     \\ %
Data to Text Generation                & \textbf{Correctness}: When  describing  predicates which  are  found  in  the  data,  does  the  text mention correct the objects and adequately introduces the subject for this specific predicate?   
                                           \\ %
                                           & \textbf{Text Structure}: Is the text grammatical, well-structured, written in acceptable English? 
                                           \\ \hline
All above tasks & \textbf{Fluency}: How fluent is the generated text? \\ \hline\hline
\end{tabular}
}
\caption{Criteria used for human judgements to evaluate NLG systems for each of the 6 tasks}
    \label{tab:NLG_tasks_criteria}
\end{table*}
\section{Criteria used in Human Evaluations}
\label{sec:human_criteria}
The goal of this work is to carefully evaluate automatic evaluation metrics with a focus on their ability to capture the diverse set of criteria used by humans while assessing NLG systems. To begin with, we describe the criteria that the output of an NLG system must satisfy for the 6 NLG tasks that we consider in this work, viz., machine translation (MT), dialog generation (DG), automatic summarisation (AS), question generation(QG), data-to-text generation (D2T) and image captioning (IC). %
Over the years, different works have proposed different criteria for evaluating NLG systems. %
In this work, we consider a popular set of criteria for each task as summarised in \citet{DBLP:journals/corr/abs-2008-12009_NLG_evaluation_metrics_survey} and presented in Table \ref{tab:NLG_tasks_criteria}. Given the wide variety of criteria used for each task, one obvious question to ask is whether we really need so many criteria or is a single overall score enough.%
One could argue that it is obvious from the definitions of the criteria that each of them is unique and a good score on one (say, fluency) may not necessarily imply a good score on another (say, coverage). However, we provide a quantitative argument for this by computing the correlations between human scores for different criteria as described below. %

\subsection{Correlations between different criteria}
\begin{figure}[!h]
\includegraphics[scale=0.8]{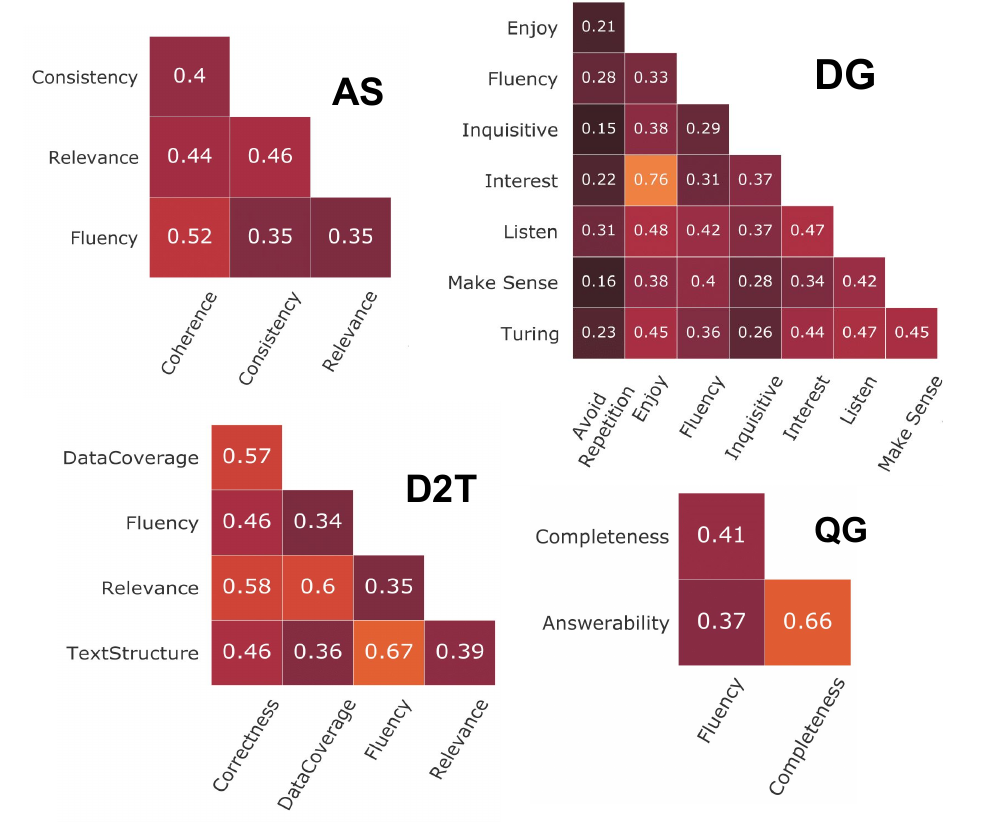}
\caption{Correlations between criteria for different tasks. \textbf{The darker the shade of the cell, the lower is the correlation}.}
\label{fig:all_criteria_correlations}

\end{figure}

We use existing publicly available datasets containing human judgement scores on multiple criteria for each of the 6 tasks described earlier. For example, \citep{castro-ferreira-etal-2020-2020_webnlg2020} contains 3025 samples of outputs generated by data-to-text generation systems that participated in the WebNLG 2020 challenge. For each of these samples, the organisers asked humans to rate the output based on 5 criteria, viz., \textit{fluency, data coverage, relevance, correctness} and \textit{text structure}. We use these scores to compute the correlations between the scores of all the $5 \choose 2$ pairs of criteria. We repeat this for the other tasks using the datasets described in Table \ref{tab:stats_datasets}\footnote{%
For AS, we could not find a dataset containing human judgements for the set of criteria in \citet{DBLP:journals/corr/abs-2008-12009_NLG_evaluation_metrics_survey}. Hence, we use the dataset provided by  \citet{DBLP:journals/corr/abs-2007-12626_summeval}.}
\footnote{Note that all of the datasets mentioned in Table \ref{tab:stats_datasets} were collected using well established methods to ensure that the annotations were of high quality. Some of these datasets do not explicitly report the Inter Annotator Agreement (IAA) scores whereas others \citep{DBLP:journals/corr/abs-2007-12626_summeval,castro-ferreira-etal-2020-2020_webnlg2020,DBLP:conf/emnlp/NemaK18_qbleu} report a good IAA score ranging from 0.63-0.71.} Using these annotations, we compute the pairwise Kendall tau correlations between all criteria for each task as seen in Figure \ref{fig:all_criteria_correlations}. (Refer to appendix \ref{sec:appendix_cri_cri_corrs} for pearson correlations of the criteria.)
\begin{table}[!t]
\resizebox{1\columnwidth}{!}{
\begin{tabular}{l|l|l}
\hline\hline
\textbf{Task} & \textbf{Data collected/ released by} & \# \textbf{Samples (annotators)} \\ \hline\hline
QG   &   \citet{DBLP:conf/emnlp/NemaK18_qbleu}      & 1000   (in-house)       \\ \hline
AS   &  \citet{DBLP:journals/corr/abs-2007-12626_summeval}       & 1600  (expert, crowdsource)        \\ \hline
D2T  &  \citet{castro-ferreira-etal-2020-2020_webnlg2020}       & 3025  (crowdsourced)        \\ \hline
DG   &  \citet{DBLP:conf/naacl/SeeRKW19}       & 3316  (crowdsourced)        \\ \hline
MT   &   \citet{callisonburch-EtAl:2007:WMT}      & 10,754  (crowdsourced)      \\ \hline
IC   &   \citet{aditya2015images_human_eval_flickrncoco} {\small (Coco subset)}     & 2007  (crowdsourced)       \\ \hline %
\end{tabular}
}
\caption{Datasets with human scores on many criteria.}
\label{tab:stats_datasets}
\end{table}

We see that, across tasks, for most pairs of criteria, the correlation is moderate (between 0.3 and 0.5) to low ($< 0.3$). The highest correlation of 0.76 is observed between \textit{interestingness} and \textit{enjoyability} for dialogue generation. %
However other criteria such as avoiding \textit{repetition}, \textit{inquisitiveness}, and \textit{making sense} have low correlations with most of the other criteria. We make similar observations for the correlations between the criteria for other tasks. Even for IC %
the correlation between the 2 criteria of thoroughness and correctness is 0.41. For MT, the commonly used criteria of \textit{fluency} and \textit{adequacy} were found to be highly correlated with Pearson correlation co-efficient of 0.69 \citep{DBLP:journals/taslp/BanchsDL15_AM_FM}. This justifies why WMT evaluations now ask humans to give to only a single score indicating overall quality. However, given the low to moderate correlations between criteria for other tasks, a similar %
strategy is not prudent for these tasks.\\ %
\textbf{Takeaway:} For tasks whose linguistic criteria show a low or at-best moderate correlation with each other, a single score assigned by a automatic metric is inadequate for a comprehensive assessment.

\section{Perturbation Checklists}
\label{sec:Perturbation_templates}
\begin{table*}[!t]
\centering %
\small
\resizebox{1\textwidth}{!}{
\rowcolors{1}{}{lightgray!50}
\begin{tabular}{l L{2cm} L{4cm} L{7.5cm} L{7.5cm}} %
\hline\hline %
\textbf{Task} & \textbf{Criteria} & \textbf{Perturbation} & \textbf{Unmodified sentence} & \textbf{Perturbed sentence} \\ %
\hline \hline%

 \cellcolor{white}   &\cellcolor{white}& Jumbling word order & We \colorbox{color}{play} badminton every evening. & We badminton every evening play. \\
    \cellcolor{white}   &\cellcolor{white} & Subject-verb disagreement   & He \colorbox{color}{doesn't} know how to bake. & He \colorbox{lightred} {don't} know how to bake.\\
   \cellcolor{white}  &\multirow{-3}{*}{\cellcolor{white}Fluency}  & Spelling errors&Make the most of every opportunity presented to you.&Make the most of \colorbox{lightred}{evry} opportunity presented to you.\\
    \hhline{~----}\hhline{~}
    \cellcolor{white} & \cellcolor{white}
    &Replace with synonyms&The mangoes are \colorbox{color}{delicious}. & The mangoes are \colorbox{lightred}{tasty}.\\ %
    \cellcolor{white} & \cellcolor{white}&Contractions&We \colorbox{color}{are} going to embark on an adventure. & \colorbox{lightred}{We're} going to embark on an adventure.\\ 
    \multirow{-6}{1.4em}{\cellcolor{white}All tasks} & \multirow{-3}{*}{\cellcolor{white}Invariance} & Numerals to words&The flight will be delayed by \colorbox{color}{2} hours.&The flight will be delayed by \colorbox{lightred}{two} hours.\\
    \cline{1-4}
    \hline
    \cellcolor{white}   &\cellcolor{white}    &Dropping out words or phrases &I \colorbox{color}{was being} followed. & I followed. \\ %
    \cellcolor{white}   &\cellcolor{white} &Add extra text& This book is so inspiring. & This book is so inspiring,\colorbox{lightred} {I forgot}. \\ 
    \multirow{-4}{1.4em}{\cellcolor{white}MT} & \multirow{-4}{*}{\cellcolor{white}Adequacy}  & Negation/antonyms & It \colorbox{color}{will rain} on Monday.&It will \colorbox{lightred}{not} rain on Monday.\\
    \hline

    \cellcolor{white} &\multirow{-2}{7em}{\cellcolor{white}Informativeness} &Use hyponyms to create misinformation& The girl my \colorbox{color}{brother} Andy met through MySpace turned out to be completely made up . & The girl my \colorbox{lightred}{friend} Andy met through MySpace turned out to be completely made up. \\ 
    \hhline{~----}\hhline{~}
    \cellcolor{white} & \cellcolor{white}Flow / coherence & Reorder sentences & The pandemic was spreading uncontrollably. Vaccines are being developed and tested rapidly.& Vaccines are being developed and tested rapidly. \colorbox{lightred}{The pandemic was spreading uncontrollably.}\\
    \hhline{~----}\hhline{~}
\multirow{-6}{*}{\cellcolor{white}AS}  &\cellcolor{white}Referential clarity& Replace nouns by pronouns& \colorbox{color}{The pandemic} was spreading uncontrollably. \colorbox{color}{Vaccines} are being developed rapidly.&\colorbox{lightred}{It} was spreading uncontrollably. \colorbox{lightred}{They} are being developed rapidly.\\
    \hline

\cellcolor{white} & \cellcolor{white}  &Drop question word& \colorbox{color}{When} was he born ? & Was he born?\\
    \cellcolor{white} & \multirow{-3}{4em}{\cellcolor{white}Answerability}   & Change question to assertive statement & Who is the director of Titanic? & \colorbox{lightred}{The director of Titanic is James Cameron.}\\
    \hhline{~----}\hhline{~}
    \cellcolor{white} & \cellcolor{white}  &Mask a few words and predict&How could Tesla run \colorbox{color}{the experiments}?&How could Tesla run to \colorbox{lightred}{the beach}?\\
 \multirow{-6}{1.4em}{\cellcolor{white}QG}    & \multirow{-2}{4em}{\cellcolor{white}Relevance} &Perturb nouns& Why did \colorbox{color}{Mary} go to the \colorbox{color}{school}? & Why did \colorbox{lightred}{Raj} go to the \colorbox{lightred}{market}? \\
 \hline

\cellcolor{white} & \cellcolor{white}
    &{\small Negate a previous statement by same speaker}& Bot: I enjoy having your daughter in my class. User: I'm glad to hear that. & I \colorbox{lightred}{don't} enjoy having your daughter in my class.\\ %
    \cellcolor{white} & \multirow{-3}{4em}{\cellcolor{white}Making sense} &Add extra non-sensible text & Do you know where Dr. XYZ lives? & Yes, \colorbox{lightred}{my father is my grandmother's father} \\
    \hhline{~----}\hhline{~}
    \cellcolor{white} & \cellcolor{white}
    & Repeat previous utterances  & Do you know Dr.XYZ?& \colorbox{lightred}{Do you know where Dr. XYZ lives?}\\
    \cellcolor{white} & \multirow{-3}{4em}{\cellcolor{white}Avoid repetition} & Repeat phrases & I like ice creams& I like ice creams, \colorbox{lightred}{ice creams}\\
    \hhline{~----}\hhline{~}
    \cellcolor{white} & \cellcolor{white}Listening & Replace with "Can you repeat?" & I need to book a taxi & \colorbox{lightred}{I'm sorry, can you repeat?}\\
    \hhline{~----}\hhline{~}
    \multirow{-9}{*}{\cellcolor{white}DG} & \cellcolor{white}Relevance & Map random responses & I am \colorbox{color}{new to coding}. & I am \colorbox{lightred}{scared of snakes}.\\
    \hline
    
    \cellcolor{white} & \cellcolor{white}
    & Change gender  &  Two \colorbox{color}{girls} are playing with a doll & Two \colorbox{lightred}{boys} are playing with a doll\\
    \cellcolor{white} & \multirow{-2}{3em}{\cellcolor{white}Correctness} & Change attributes  &  A \colorbox{color}{small} boy playing with a \colorbox{color}{red} ball & A \colorbox{lightred}{tall} boy playing with a \colorbox{lightred}{green} ball\\
    \hhline{~----}\hhline{~}
    \cellcolor{white} & \cellcolor{white}
    & Drop objects/noun  & A small \colorbox{color}{boy} playing with a red \colorbox{color}{ball} &  A small playing with a red \\
    \multirow{-5}{*}{\cellcolor{white}IC} & \multirow{-2}{4em}{\cellcolor{white}Thoroughness} & Repeat (append) object & A lady riding a horse. & A lady riding a horse \colorbox{lightred}{and a lady}.\\
    \hline

    \cellcolor{white} & \cellcolor{white} & Use hyponyms & Beethoven was a German \colorbox{color}{musician} & Beethoven was a German \colorbox{lightred}{architect} \\
    \cellcolor{white} & \multirow{-2}{4em}{\cellcolor{white}Correctness} & Change numbers  &  The cricketer was born in \colorbox{color}{1990}. & The cricketer was born in \colorbox{lightred}{1950}.\\
    \hhline{~----}\hhline{~}
    \cellcolor{white} & \cellcolor{white}
    & Drop phrases  & A small boy playing with a \colorbox{color}{red ball} &  A small boy playing with a \\
    \cellcolor{white} & \multirow{-2}{4em}{\cellcolor{white}Data Coverage} & Repeat phrases & Beethoven was a German musician & Beethoven was a German musician \colorbox{lightred}{and German musician}.\\
    \hhline{~----}\hhline{~}
    \multirow{-5}{*}{\cellcolor{white}D2T}& \cellcolor{white}Relevance & Perturb names & \colorbox{color}{Phillips} was a child prodigy. & \colorbox{lightred}{James} was a child prodigy.\\
    \hline

\hline
\end{tabular}}
\caption{Perturbation templates targeting various criteria with examples. The blue highlights indicate the portions of the original sentence affected by the perturbation template. The red highlights indicate the changes in the modified sentence.}
\label{table:examples} %
\end{table*}
So far we have established that if automatic evaluation metrics are to be used as a substitute for human evaluations as a whole, 
then they should be capable of evaluating the output on multiple desired criteria. However, the current recipe of proposing and evaluating evaluation metrics does not take this into account. 
To enable such a systematic evaluation of automatic evaluation metrics, we propose \textit{perturbation checklists}. Similar to the original Checklist paper \cite{DBLP:conf/acl/RibeiroWGS20_checklist}, the idea is to evaluate the performance of the evaluation metric in detecting criteria-specific changes in the output.

We design such perturbation templates for each relevant criteria for each of the 6 tasks as shown in Table \ref{table:examples}. For example, consider the criteria \textit{fluency} which is relevant for all the tasks. Now consider a perturbation template for this criteria which simply drops the stop words in the output. Such a perturbation would definitely affect the fluency of the output. %
If an automatic evaluation metric is capable of assessing fluency, then this drop in the fluency of the output should get reflected in the score assigned by the metric. More formally, let $p$ be the original output and $\hat{p}^{t}_{c}$ be the output obtained by applying the perturbation template $t$ for the criteria $c$. Further, let $f_e(p)$ be the score assigned by a given evaluation metric $e$ to the output $p$, normalised to be in the range $[0,1]$. If the metric $e$ is capable of assessing fluency then we would expect $f_e(\hat{p}^{t}_{c})$ to be lower than $f_e(p)$. Now, further let $h(p)$ and $h(\hat{p}^{t}_{c})$ be the scores (also normalised to have range $[0,1]$) assigned to the original and perturbed outputs by human annotators. We then define a metric $s^t_c(e)$ which captures the ability of the metric $e$ to detect the perturbation $t$ for the desired criteria $c$.
\begin{equation}
\label{eq:deviation}
    s^t_c(e) = (h(\hat{p}^{t}_{c}) - h(p)) - (f_e(\hat{p}^{t}_{c}) - f_e(p))
\end{equation}
The score $s^t_c(e)$ as defined above thus captures the deviation between a human's perception about the drop in the quality and the metric $e$'s perception about the drop in the quality. 

\definecolor{blue1}{rgb}{0.0, 0.40, 0.95}
\tikzstyle{abstract}=[rectangle, draw=black, rounded corners, fill=white, drop shadow,
        text centered, anchor=north, text=black, text width=3cm]
\tikzstyle{comment}=[rectangle, draw=black, rounded corners, fill=white, drop shadow,
        text centered, anchor=north, text=black, text width=3cm]
\tikzstyle{myarrow}=[->, >=open triangle 90, thick]
\tikzstyle{line}=[-, thick]

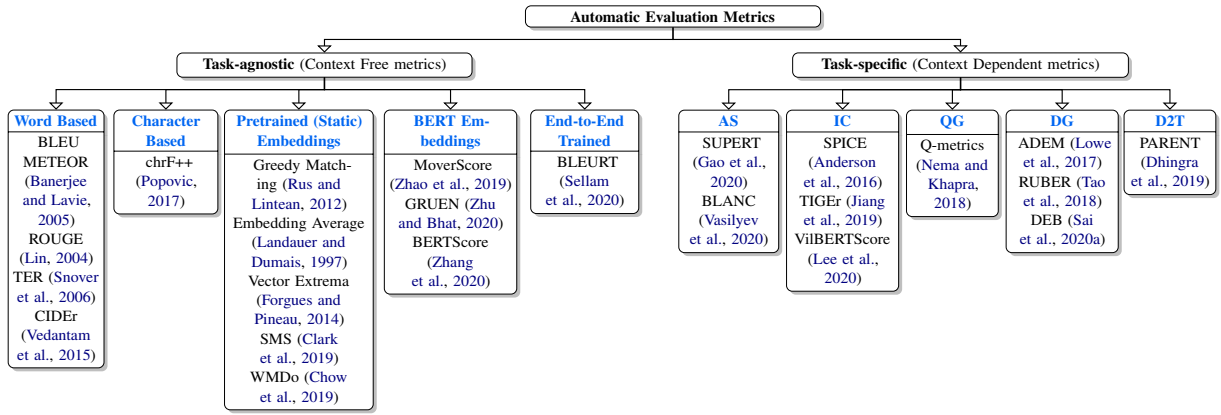
\begin{figure*}
\resizebox{1\textwidth}{!}{
\begin{tikzpicture}[node distance=2.8cm]
    \node (Root) [abstract, text width=7cm]
        {
            \textbf{Automatic Evaluation Metrics}
        };
    \node (AuxNode01) [text width=4cm, below=of Root] {};
    \node (Context_Free) [abstract, text width=7cm, left=of AuxNode01,yshift=2.0cm]
        {
            \textbf{Task-agnostic} (Context Free metrics)
        };
    \node (Context_Dependent) [abstract, text width=8cm, right=of AuxNode01,yshift=2.0cm ,xshift=-2cm]
        {
            \textbf{Task-specific} (Context Dependent metrics)
        };
        
    \node (AuxNode03) [below=of Context_Free] {};
    \node (Word) [abstract, text width=2.1cm, rectangle split, rectangle split parts=2, below=of Context_Free, xshift=-6.6cm,yshift=2cm]
        {
            {\textcolor{blue1}{\textbf{Word Based}}} \\
            \nodepart{second}{
            BLEU   \\
            METEOR \cite{meteor} \\
            ROUGE \cite{rouge} \\
            TER \cite{Snover06astudy_ter}\\
            CIDEr \cite{DBLP:conf/cvpr/VedantamZP15_CIDEr}
            }
        };
    \node (Character) [abstract, text width=2.3cm, rectangle split, rectangle split parts=2, below=of Context_Free, xshift=-3.9cm,yshift=2cm]
        {
            {\textcolor{blue1}{\textbf{Character Based}}} \\
            \nodepart{second}{
            chrF++ \cite{DBLP:conf/wmt/Popovic17_chrFpp}
            }
        };
    \node (Embedding) [abstract, text width=3.4cm, rectangle split, rectangle split parts=2, below=of Context_Free, xshift=-0.63cm,yshift=2cm]
        {   
            {\textcolor{blue1}{\textbf{Pretrained (Static) Embeddings}}} \\
            \nodepart{second}{
            Greedy Matching \cite{greedymatch} \\
            Embedding Average \cite{landauer1997solution_vector_averaging1} \\
            Vector Extrema \cite{vectorextrema} \\
            SMS \cite{DBLP:conf/acl/ClarkCS19_sms} \\
            WMDo  \cite{chow-etal-2019-WMDO} %
            }
        };

    \node (Feature_based) [abstract, rectangle split, rectangle split parts=2, below=of Context_Free, xshift=3.1cm,yshift=2cm]
        {
            {\textcolor{blue1}{\textbf{BERT Embeddings}}}\\ %
            \nodepart{second}{
            MoverScore \cite{DBLP:conf/emnlp/ZhaoPLGME19_MoverScore}\\
            GRUEN \cite{DBLP:conf/emnlp/ZhuB20_gruen}\\
            BERTScore \cite{DBLP:conf/iclr/ZhangKWWA20_BERTscore}
            }
        };
    \node (Hypothesis_based) [abstract, rectangle split, rectangle split parts=2, text width=2.4cm, below=of Context_Free, xshift=6.4cm,yshift=2cm]
    {
        {\textcolor{blue1}{\textbf{End-to-End Trained}}}\\
        \nodepart{second}{
        BLEURT \cite{DBLP:conf/acl/SellamDP20_bleurt}
        }
    };
    \node (AuxNode05) [below=of Context_Dependent] {};
        
    \node (AS) [abstract, text width= 2.2cm, rectangle split, rectangle split parts=2, below=of Context_Dependent, xshift=-5.7cm,yshift=2cm]
        {
            {\textcolor{blue1}{\textbf{AS}}}\\
            \nodepart{second}{
            SUPERT \cite{DBLP:conf/acl/GaoZE20_supert} \\
            BLANC \cite{DBLP:journals/corr/abs-2002-09836_blanc}
            }
        };
    \node (IC) [abstract, text width= 2.5cm, rectangle split, rectangle split parts=2, below=of Context_Dependent, xshift=-2.9cm,yshift=2cm]
        {
            {\textcolor{blue1}{\textbf{IC}}}\\
            \nodepart{second}{
            SPICE \cite{DBLP:conf/eccv/AndersonFJG16_SPICE}\\
            TIGEr \cite{DBLP:conf/emnlp/JiangHZWZGDG19_TIGEr}\\
            VilBERTScore \cite{lee-etal-2020-vilbertscore}
            }
        };    
     \node (AD) [abstract, text width= 2cm, rectangle split, rectangle split parts=2, below=of Context_Dependent, xshift=-0.2cm,yshift=2cm]
        {
        {\textcolor{blue1}{\textbf{QG}}}\\
        \nodepart{second}Q-metrics \cite{DBLP:conf/emnlp/NemaK18_qbleu}
        };    
    \node (QA) [abstract, text width= 2.5cm, rectangle split, rectangle split parts=2, below=of Context_Dependent, xshift=2.5cm,yshift=2cm]
        {
            {\textcolor{blue1}{\textbf{DG}}}\\
            \nodepart{second}{
            ADEM \cite{DBLP:conf/acl/LoweNSABP17ADEM}\\
            RUBER \cite{DBLP:conf/aaai/TaoMZY18ruber}\\
            DEB \cite{DBLP:journals/tacl/SaiMAK20_DDpp}
            }
        };    
    \node (D2T) [abstract, text width= 2cm, rectangle split, rectangle split parts=2, below=of Context_Dependent, xshift=5.15cm,yshift=2cm]
        {
            {\textcolor{blue1}{\textbf{D2T}}}\\
            \nodepart{second}PARENT \cite{DBLP:conf/acl/DhingraFPCDC19_parent}
        };

    \draw[myarrow] (Root.south) -- ++(0,-0.3) -| (Context_Free.north);
    \draw[myarrow] (Root.south) -- ++(0,-0.3) -| (Context_Dependent.north);

    \draw[myarrow] (Context_Free.south) -- ++(0,-0.3) -| (Word.north);
    \draw[myarrow] (Context_Free.south) -- ++(0,-0.3) -| (Character.north);
    \draw[myarrow] (Context_Free.south) -- ++(0,-0.3) -| (Embedding.north);

    \draw[myarrow] (Context_Free.south) -- ++(0,-0.3) -| (Feature_based.north);
    \draw[myarrow] (Context_Free.south) -- ++(0,-0.3) -| (Hypothesis_based.north);
     
    \draw[myarrow] (Context_Dependent.south) -- ++(0,-0.3) -| (AS.north);
    \draw[myarrow] (Context_Dependent.south) -- ++(0,-0.3) -| (IC.north);
    \draw[myarrow] (Context_Dependent.south) -- ++(0,-0.3) -| (QA.north);
    \draw[myarrow] (Context_Dependent.south) -- ++(0,-0.3) -| (D2T.north);
    \draw[myarrow] (Context_Dependent.south) -- ++(0,-0.3) -| (AD.north);

\end{tikzpicture}
}
\caption{Metrics analysed in this study }%
\label{fig:metrics_used_taxo_fig}
\end{figure*}

We design a total of 34 such perturbation templates across all the criteria and all the tasks. Each template is manually created by us and targets a specific criteria. We also present \textit{invariant} templates that do not affect any criteria although they modify the sentences. For perturbations resulting from such invariant templates the score of the metric should not drop. The invariant and fluency-based templates are common for all the tasks considered in this work. Table \ref{table:examples} shows sample perturbations generated by each of the templates. (Please refer to appendix \ref{sec:appendix_full_templates} for a more comprehensive list of the proposed perturbations with examples for each task). These perturbed sentences are generated automatically using the checklist framework \cite{DBLP:conf/acl/RibeiroWGS20_checklist}. This framework contains modules for performing simple string manipulations such as dropping stop words, replacing/dropping named entities, masking words or replacing them by other words/phrases. %
We also extend the framework with additional modules for jumbling words, changing numbers to words, subject-verb disagreement, changing gender, reordering sentences, adding spurious text, and adding redundancy at the word/ phrase /sentence-level.%

\section{Experimental setup}
\label{sec:experimental_setup}
We first do a coarse grained evaluation of several metrics by computing their correlations with the scores assigned by humans for multiple criteria. 
Note that unlike existing studies which study such correlations for a small number of metrics (typically, $n$-gram based metrics) for a specific task (say, MT) and a single criteria (typically, overall quality), we do a more comprehensive study involving a combination of 6 tasks, 25 metrics and multiple criteria. Apart from this coarse grained evaluation which simply looks at correlations, we also do a more fine-grained evaluation of the robustness of these metrics to different criteria-specific perturbations as summarised in Table \ref{table:examples}. This fine-grained evaluation augments the coarse-grained evaluation and helps us understand the evaluation capabilities of these metrics. Below, we describe the datasets and evaluation metrics used in our work.

\begin{figure*}[!ht]
\centering
\begin{subfigure}{0.49\textwidth}
    \centering
    \includegraphics[width=1\columnwidth]{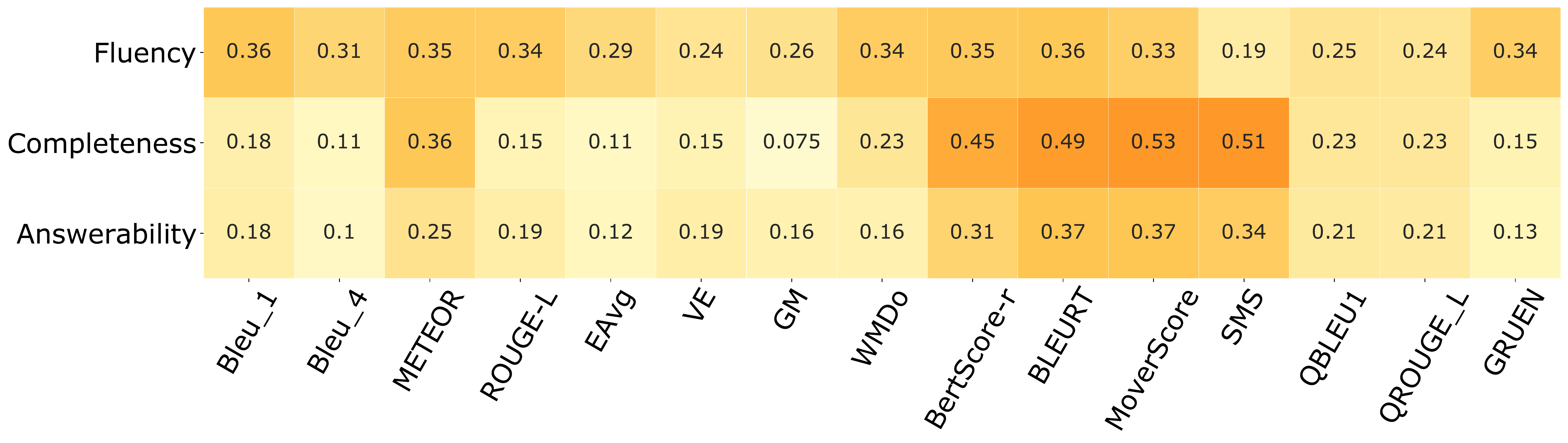}
    \vspace{-8mm}
    \caption{Question Generation}
    \label{fig:qg_met_corrs}
    \centering
    \includegraphics[width=1\columnwidth]{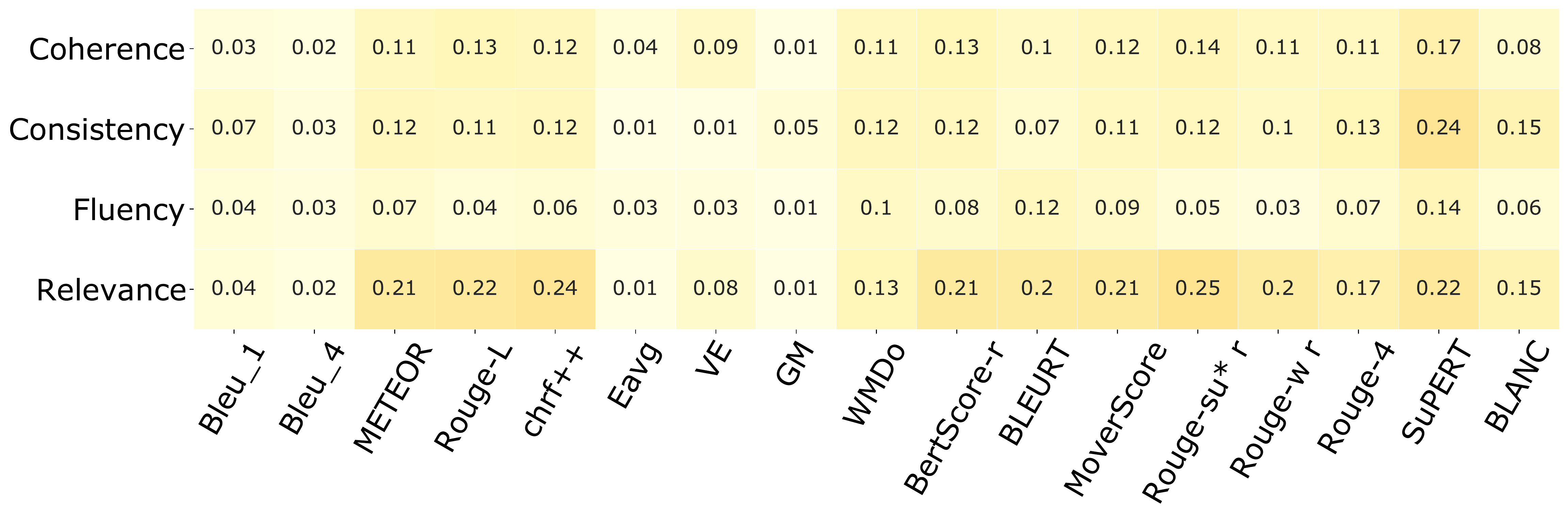}
    \vspace{-8mm}
    \caption{Abstractive Summarization}
    \label{fig:as_met_corrs}
    \centering
    \includegraphics[width=1\columnwidth]{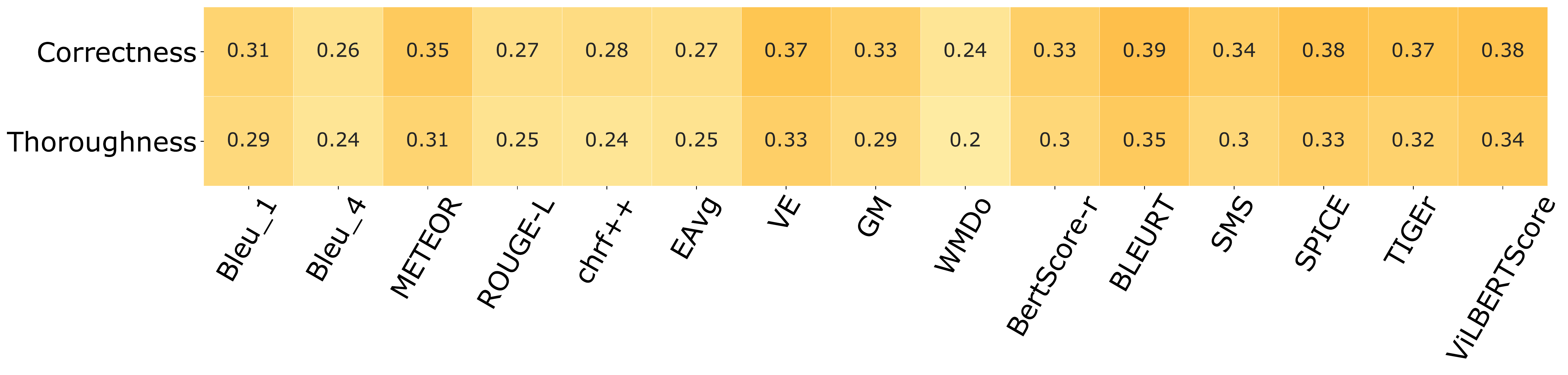}
    \vspace{-8mm}
    \caption{Image Captioning}
    \label{fig:ic_met_corrs}
\end{subfigure}
\begin{subfigure}{0.49\textwidth} 
    \centering
    \includegraphics[width=1\columnwidth]{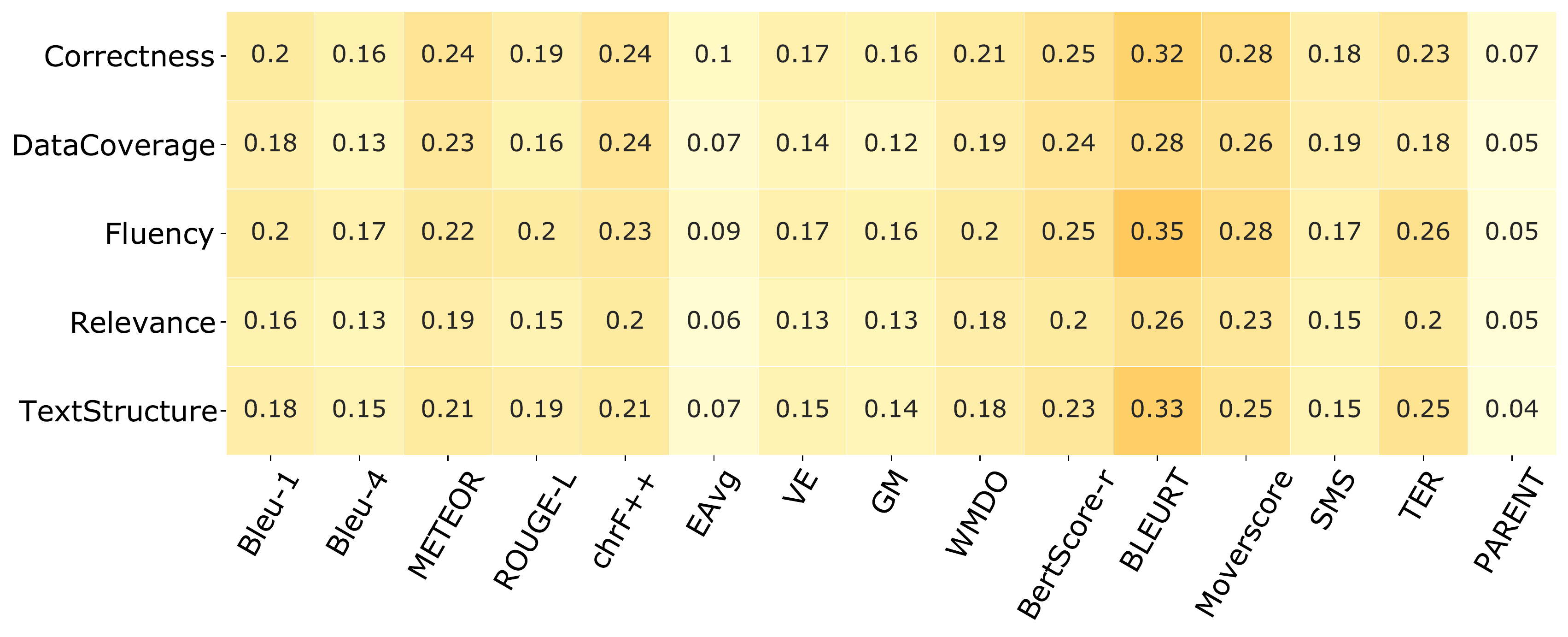}
    \vspace{-8mm}
    \caption{Data-to-Text Generation}
    \label{fig:dt_met_corrs}
    \centering
    \includegraphics[width=1\columnwidth]{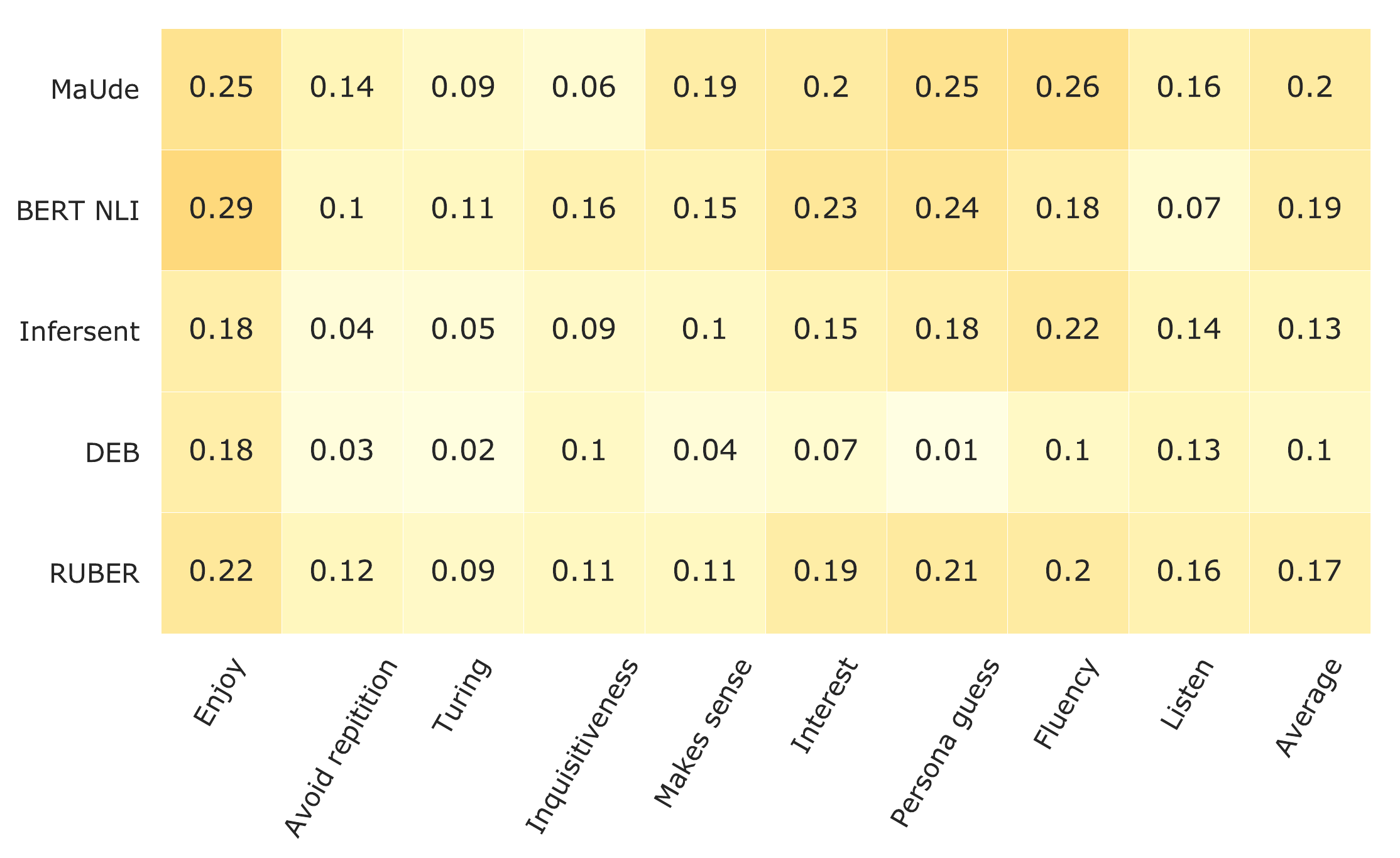}
    \vspace{-8mm}
    \caption{Dialogue Generation}
    \label{fig:dg_met_corrs}
\end{subfigure}
\caption{Correlations of metrics with different criteria (Note: For MT, we refer to the results of the WMT 2020 shared metrics task \cite{DBLP:conf/wmt/MathurWFMB20_wmt-20-results}. For DG, we only present the reference-less metrics since the dataset does not contain references to compute the task-agnostic metric scores.) (Refer appendix \ref{sec:appendix_full_plots} for full plots) }%
\label{fig:all_metric_criteria_correlations}
\end{figure*}

\noindent \textbf{Datasets}. For the coarse-grained evaluation, we use the datasets containing human judgements as described earlier in Table \ref{tab:stats_datasets} in Section \ref{sec:human_criteria}. For the fine-grained evaluation, we use datasets containing multiple ground truth references which can then be perturbed using our templates. %
For MT, we use the expanded version of newstest2017 Chinese to English dataset~\cite{DBLP:journals/corr/abs-1803-05567_MT_extra_references} which contains two references for each sentence. For QG, we use the SQuAD dataset \cite{DBLP:conf/emnlp/RajpurkarZLL16_squad_dataset} which contains multiple questions for each passage. For AS, we use the curated personal narrative corpus \cite{DBLP:conf/eacl/McKeownOC17_AS_reddit_personal_reddit}. %
For DG, we use DailyDialog++ \cite{DBLP:journals/tacl/SaiMAK20_DDpp} which contains two-speaker conversations on generic topics. For IC, we use the COCO component of the Composite dataset containing 5 reference captions for each image \cite{aditya2015images_human_eval_flickrncoco}. Lastly, for D2T, we use the Triples-to-Text data of the WebNLG 2020 challenge dataset \cite{castro-ferreira-etal-2020-2020_webnlg2020}.

\noindent \textbf{Applying perturbations}. We take the reference sentences from the above task-specific datasets and apply perturbations using the Checklist framework described earlier in section \ref{sec:Perturbation_templates}. %
We first preprocess the sentences by performing tokenization, part-of-speech tagging, named entity recognition, \textit{etc}. The targeted part of the sentence is then modified either by leveraging simple string manipulation functions or by masking and generating the words/ phrases using the predictions by RoBERTa~\cite{DBLP:journals/corr/abs-1907-11692_roberta}. We provide more implementation details in appendix \ref{sec:appendix_perturbation_details}. %

\noindent \textbf{Automatic Evaluation Metrics.}
We study a total of 25 evaluation metrics belonging to different classes as shown in Figure~\ref{fig:metrics_used_taxo_fig}. %
For BLEU \cite{bleu}, METEOR \cite{meteor}, ROUGE-L \cite{rouge}, CIDEr \cite{DBLP:conf/cvpr/VedantamZP15_CIDEr}, Embedding Averaging (EAvg) \cite{landauer1997solution_vector_averaging1}, Greedy Matching (GM) \cite{greedymatch}, and Vector Extrema (VE) \cite{vectorextrema}, we use the implementation provided by \citet{sharma2017nlgeval}. For chrF++ \cite{DBLP:conf/wmt/Popovic17_chrFpp}, TER \cite{Snover06astudy_ter}, BERTScore \cite{DBLP:conf/iclr/ZhangKWWA20_BERTscore}, and BLEURT \cite{DBLP:conf/acl/SellamDP20_bleurt} we use the repository of \citet{castro-ferreira-etal-2020-2020_webnlg2020}. For SMS \cite{DBLP:conf/acl/ClarkCS19_sms}, WMDo \cite{chow-etal-2019-WMDO}, and MoverScore \cite{DBLP:conf/emnlp/ZhaoPLGME19_MoverScore}, we use the implementation provided by \citet{DBLP:journals/corr/abs-2007-12626_summeval}. For all the task-specific metrics in Figure~\ref{fig:metrics_used_taxo_fig}, we use the official codes from the respective papers.

\noindent \textbf{Collecting human judgements.} For \textit{eq}~\ref{eq:deviation}, we need human judgement scores. We collect these with the help of 15 annotators who were computer science graduates with a background in the field of Natural Language Processing (NLP) and are also proficient in English language. For each task and criteria, the annotators were provided with the corresponding perturbation templates and asked to provide a penalty score indicating by how much a perturbation would alter the meaning/essence of a sentence on a scale of 0-10. A score of 0 indicates that there is no difference between the original and modified sentences upon application of the perturbation whereas a score of 10 indicates that the perturbation drastically alters sentences. These scores are normalised to be in the range $[0,1]$. The term $h(\hat{p}^t_c)$ in \textit{eq}~\ref{eq:deviation} for perturbation $t$ is computed by subtracting the mean of all normalised human penalty scores from $1$. 

The standard deviation of the normalised annotator scores for each perturbation lies between 0.03 to 0.2. We also measure the inter annotator agreement by splitting annotators randomly into 2 groups and computing the kendall tau correlation score between the average scores of the 2 groups following \citet{DBLP:conf/emnlp/LiuLSNCP16HowNot}. This process was run 5 times with different seeds and the final inter-annotator correlation score was found to be 0.79. %

\begin{figure*}[!ht]
\begin{subfigure}{.49\textwidth}
    \centering
    \includegraphics[width=1\columnwidth]{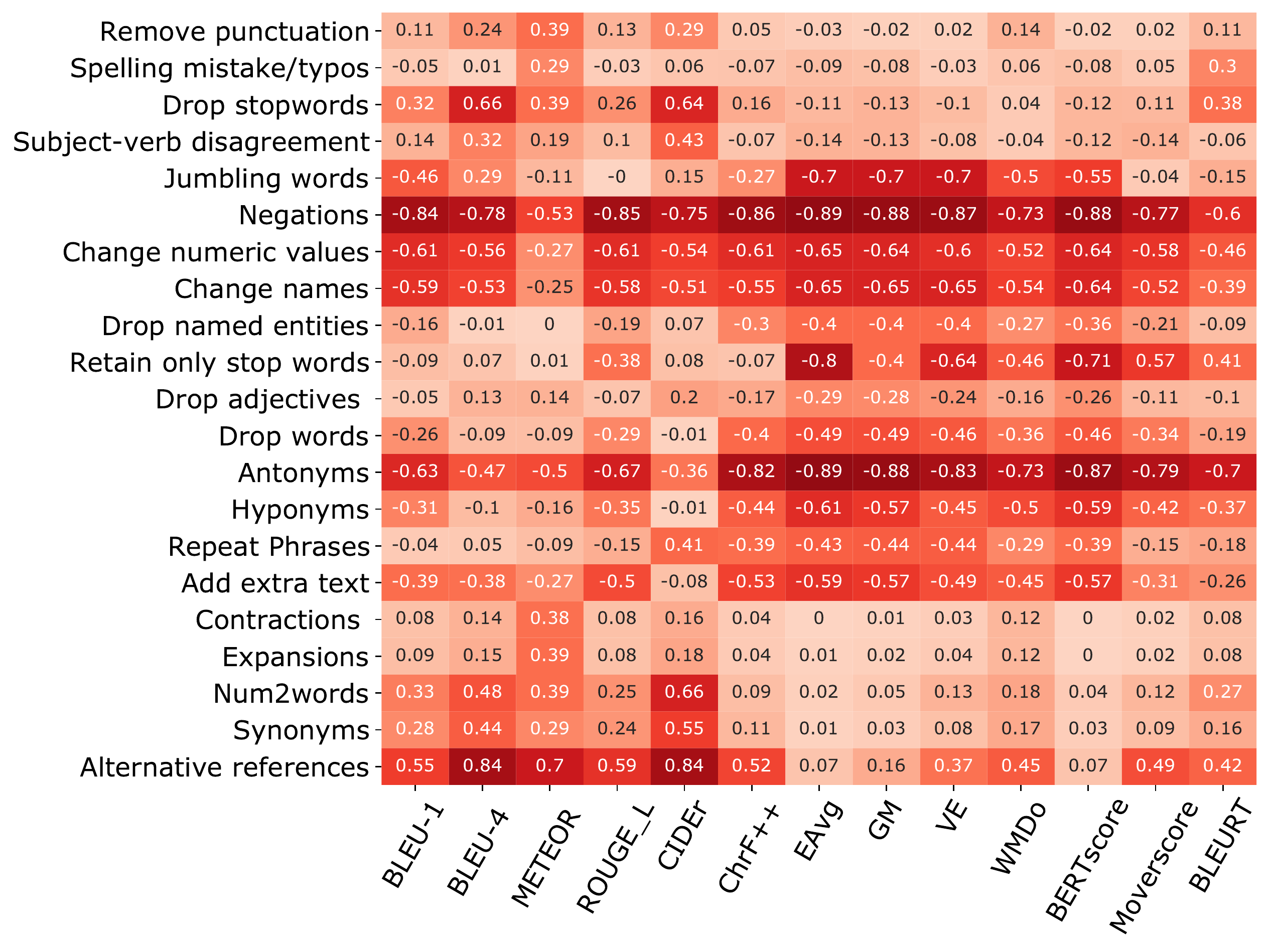}
    \vspace{-8mm}
    \caption{Machine Translation}
    \label{fig:div_mt_heatmap}
\end{subfigure}
\begin{subfigure}{.49\textwidth}
    \centering
    \includegraphics[width=1\columnwidth]{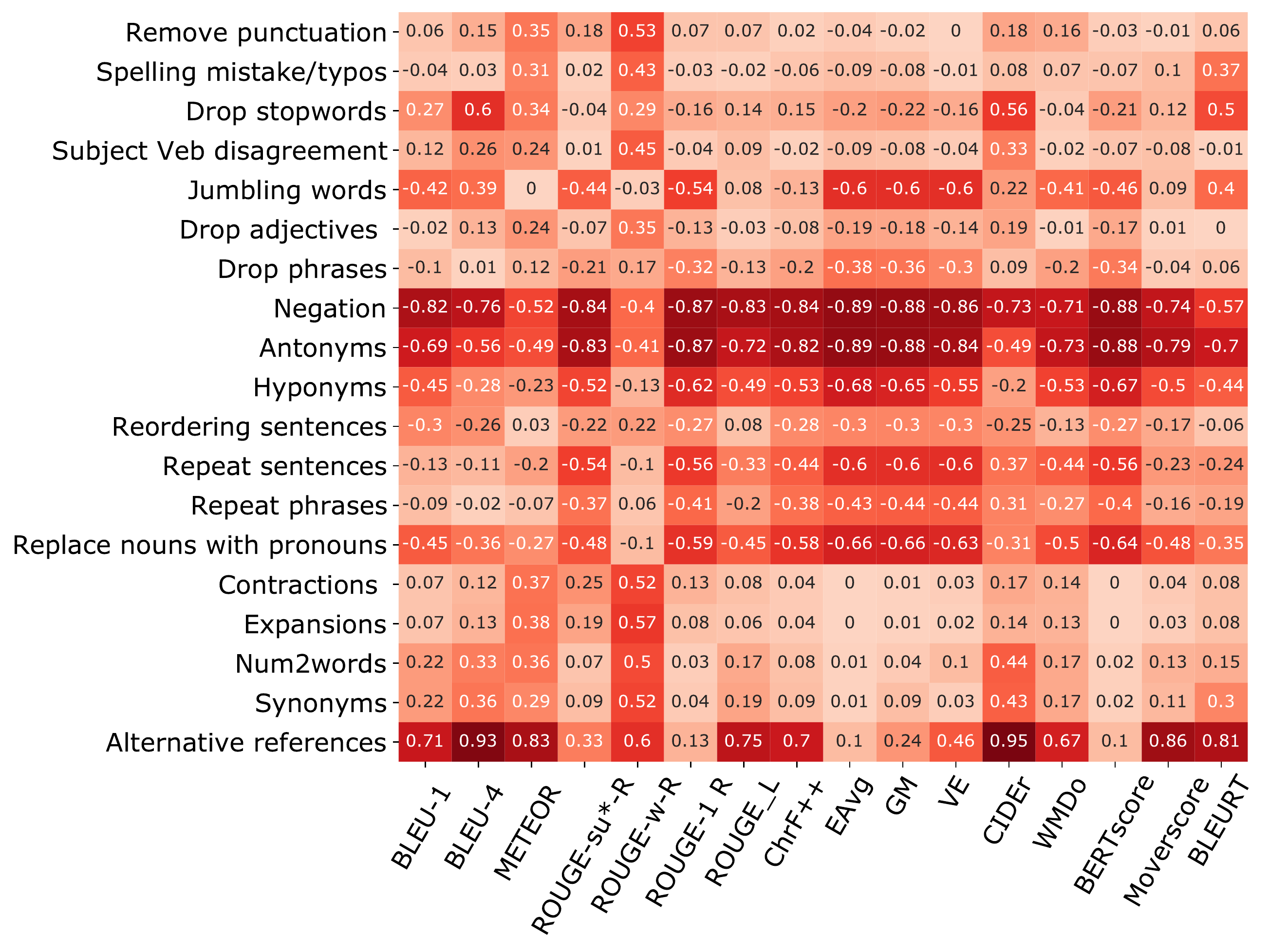}
    \vspace{-8mm}
    \caption{Abstractive Summarization}
    \label{fig:div_as_heatmap}
\end{subfigure}
\begin{subfigure}{.49\textwidth}
    \centering
    \includegraphics[width=1\columnwidth]{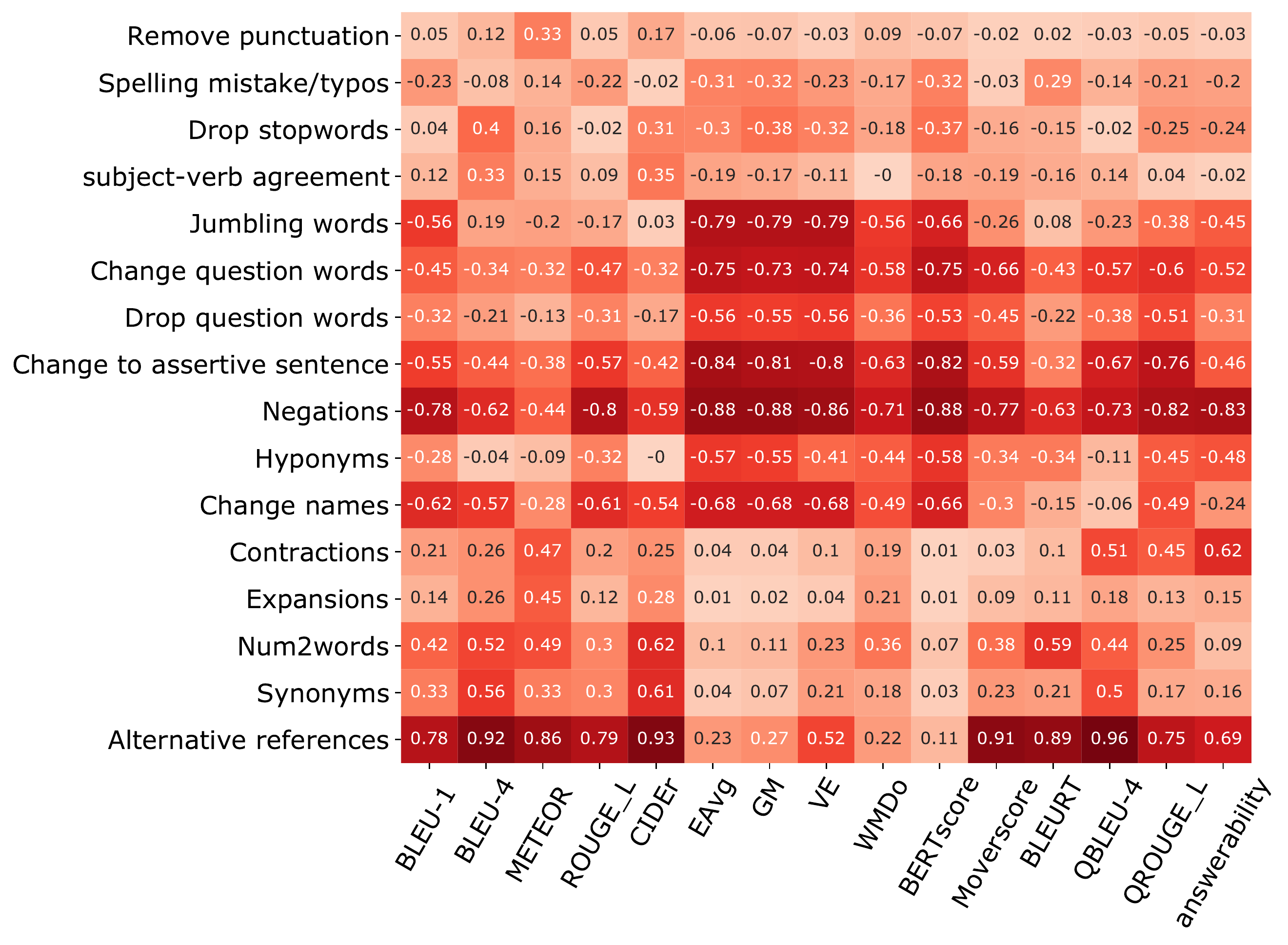}
    \vspace{-8mm}
    \caption{Question Generation}
    \label{fig:div_qg_heatmap}
\end{subfigure}
\begin{subfigure}{.49\textwidth}
    \centering
    \includegraphics[width=1\columnwidth]{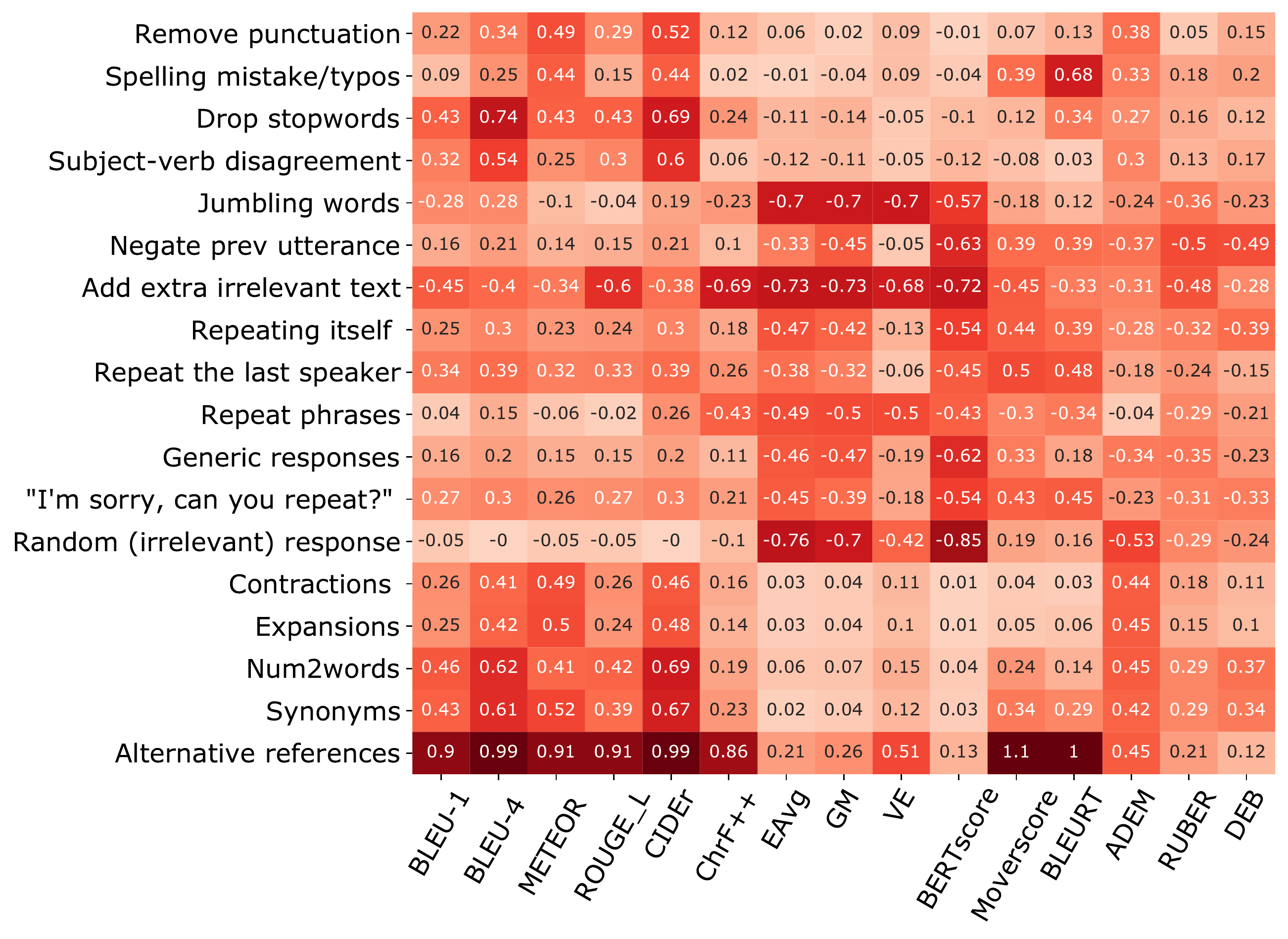}
    \vspace{-8mm}
    \caption{Dialogue Generation}
    \label{fig:div_dg_heatmap}
\end{subfigure}
\begin{subfigure}{.49\textwidth}
    \centering
    \includegraphics[width=1\columnwidth]{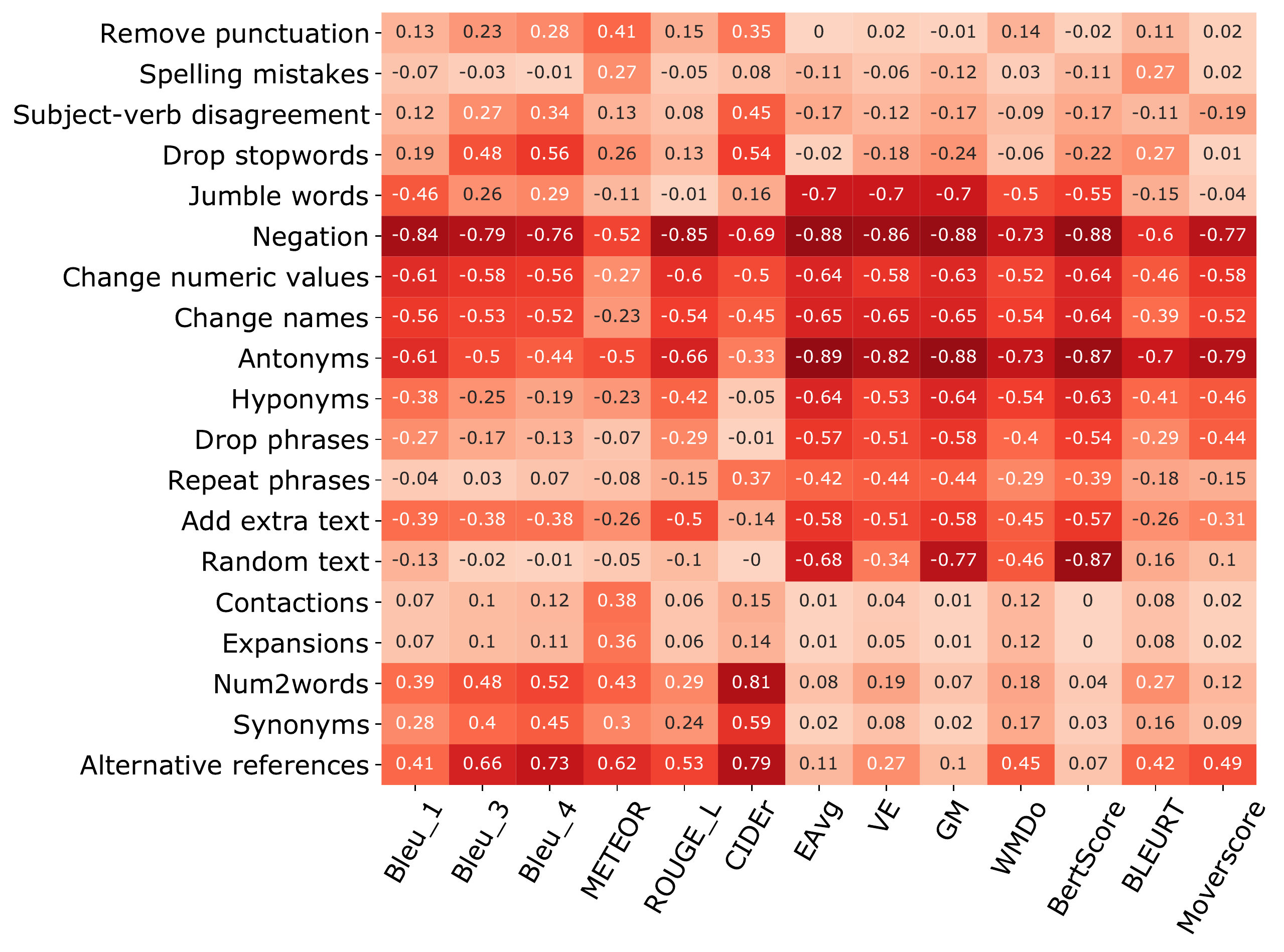}
    \vspace{-8mm}
    \caption{Data-to-Text Generation}
    \label{fig:div_dt_heatmap}
\end{subfigure}
\begin{subfigure}{.49\textwidth}
    \centering
    \includegraphics[width=1\columnwidth]{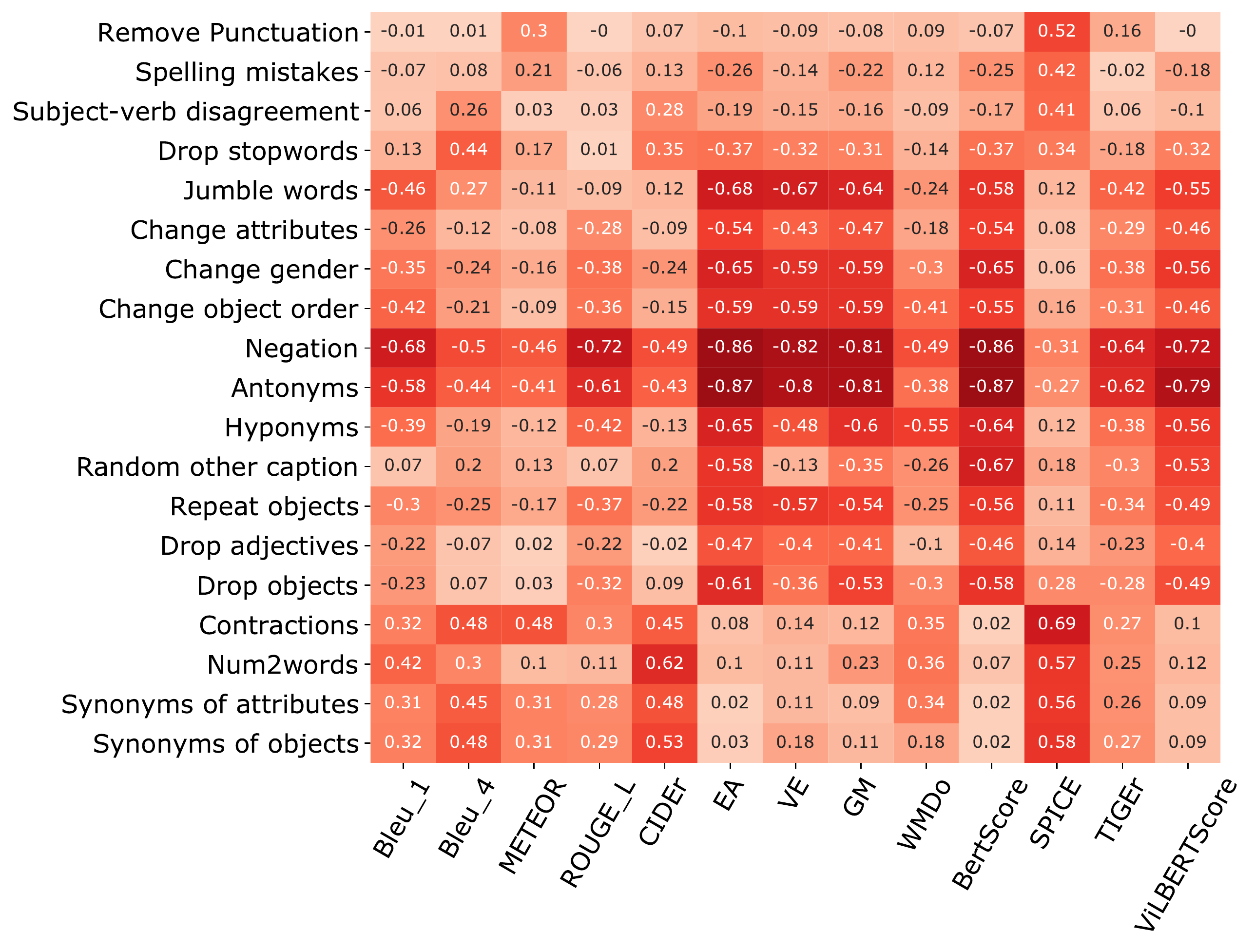}
    \vspace{-9mm}
    \caption{Image Captioning}
    \label{fig:div_dicheatmap}
\end{subfigure}
\caption{Heatmap of deviation of metric scores from human averages. The darker the cell, the more the absolute deviation from human scores.}
\label{fig:div_heatmaps}
\end{figure*}

\section{Results and Discussion}
We now discuss the results of our experiments. 

\subsection{Insights from coarse-grained evaluation}
Figure \ref{fig:all_metric_criteria_correlations} shows the correlations of different automatic evaluation metrics with multiple evaluation criteria, across 6 different NLG tasks. Our main observations from the figure are: 

\noindent \textbf{Mostly poor correlations of metrics across criteria.} We observed that across all tasks and all criteria, most of the metrics have poor correlations. In particular, out of the 271 correlation values reported in Figure \ref{fig:all_metric_criteria_correlations}, 228 are poor ($<$0.3), 35 are moderate (between 0.3 and 0.5) and only 2 are high ($>$0.5). Surprisingly even some very recently proposed metrics such as DEB \cite{DBLP:journals/tacl/SaiMAK20_DDpp}, BLANC \cite{DBLP:journals/corr/abs-2002-09836_blanc} and MaUde \cite{DBLP:conf/acl/SinhaPWLHP20_maude} do not correlate well with human judgements on other criteria. This is despite the fact that these are task-specific metrics which use the modern machinery of pre-trained BERT-based models and are fine-tuned on human judgements for overall quality. This vindicates our stand that simply tuning for overall quality does not lead to good correlations with other criteria. We do observe a few decently correlated metrics for some of the tasks along a few criteria. Specifically, the moderate correlations are found (i) in D2T for BLEURT along the dimensions of fluency, correctness and text structure, (ii) in IC for all of the task-specific metrics, majority of the embedding-based metrics and for METEOR, (iii) in QG for most metrics (except SMS) along fluency, for BERTScore, Moverscore, SMS and BLEURT along answerability, as well as along completeness, on which we find some of the highest correlations. %

\noindent \textbf{Pre-training and/or training often helps.} While almost all the metrics have poor correlations with different criteria across tasks, we observe that the ones which use a pre-trained component such as static or contextualised word/sentence embeddings and/or use task-specific training data perform better. For example, BLEURT which uses a pre-trained BERT and is fine-tuned using human judgements for MT, is among the top performing metrics across all the tasks. These findings are also consistent with those reported in the WMT20 shared task on evaluation metrics for MT \cite{DBLP:conf/wmt/MathurWFMB20_wmt-20-results}. 

\noindent \textbf{Task-agnostic metrics versus task-specific metrics.} For the tasks of QG, AS, IC and D2T we find that task-agnostic metrics such as BERTScore, Moverscore and BLEURT are consistently among the top performing metrics (i.e., their correlation scores are either the best or close to the best scores for a given task and criteria) along with the task-specific metrics. This is interesting as these task-agnostic metrics were not fine-tuned on any task specific human judgements and were originally proposed for a different task (MT). %
Among the task-specific metrics, SUPERT for AS has a relatively better correlation with consistency than all other metrics. Overall, there seems to be scope for more work/improvements on task-specific metrics to capture the criteria peculiar to each task. %

\subsection{Insights from fine-grained evaluation}
We now complement the above analysis with a more fine-grained analysis using perturbation templates. %
To do so, we use the perturbation templates described in Section \ref{sec:Perturbation_templates} and plot the deviation between metric scores and human scores using the formula in Equation \ref{eq:deviation}. These results are presented in Figure \ref{fig:div_heatmaps} and summarised below.

\noindent\textbf{Correlations do not reveal everything.} In the previous section we observed that BERT-based metrics such as BERTScore, BLEURT and MoverScore are among the top performing metrics across tasks and criteria. However, our anslysis with perturbation templates reveals that even these metrics are not robust to very simple perturbations. For example, for the task of MT consider the perturbations of adding negations, changing names, changing numeric values or replacing by antonyms in the output which can significantly alter the meaning of the sentence and thereby affect adequacy. However, BERTScore, BLEURT and MoverScore are not able to perceive this drop in quality and have a substantial deviation from human scores. We make similar observations across tasks that such metrics are not able to detect these simple perturbations. %

\noindent\textbf{Task-specific nuances are not captured.} Our analysis also shows that existing metrics are not capable of addressing well known task specific grievances. For example, for the task of DG, it is known that many NLG systems generate generic responses which leads to poor engagement with the users. However, none of the metrics are sensitive to perturbations producing `generic responses' such as \textit{ok}, \textit{thanks} or back-off responses such as \textit{I`m sorry, can you repeat?}\footnote{A model that frequently produces generic responses is undesirable. Hence, the expert human annotators assigned a high penalty to this perturbation. Most of the task-specific metrics do use the context as an input whereas, the task-agnostic metrics do not.} Similarly, for the task of QG, \citet{DBLP:conf/emnlp/NemaK18_qbleu} show that the \textit{answerability} of a question is affected if we drop/replace question words or change named entities in the question. However, we find that most metrics (including  the task specific QBLEU4/QROUGE) are not sensitive to such perturbations with very high deviation from human scores. On similar lines, ViLBERTScore which is a state of the art evaluation metric for IC is not sensitive to perturbations in gender, order of objects or attributes used for describing objects. This is of concern as many IC systems are known to produce generic captions containing genders, attributes and objects which are most prevalent in the training data. Similarly, for the task of D2T, where coverage and factual correctness are important we observe that most metrics are unable to detect perturbations which add extra/random text to the output or drop named entities (which often contain the most important information). Lastly, for AS it is important that an evaluation metric should penalise summaries which are not coherent or contain redundant sentences or do not have referential clarity. However, we observe that most metrics are not sensitive to perturbations which reorder the sentences or repeat sentences/phrases or replace nouns with pronouns (affecting referential clarity). %

\noindent\textbf{Different metrics have different skills.} While no single metric is capable of detecting all types of perturbations, we observe that some metrics are more robust to certain perturbations. BLEURT and Moverscore are robust to jumbling of words, but BERTScore is not, revealing their differences in detecting \textit{fluency}. Moverscore, BERTScore and the embedding-based metrics like Greedy Matching and Embedding Average are quite robust to simple transformations of converting numbers to corresponding words, which is an important criteria for the task of D2T, while BLEURT is relatively less robust to it. Similarly, while BERTScore performs poorly for many perturbations, it is able to respect alternative references, i.e., similar to humans, it does not drop its score when presented with alternative correct references from the dataset (last row in Figure \ref{fig:div_mt_heatmap} to \ref{fig:div_dicheatmap}). An interesting observation from the IC task is that SoTA metrics like SPICE and ViLBERTScore show a complementary behaviour on our set of perturbation criteria (third to last and last column in Figure \ref{fig:div_dicheatmap}). This opens up interesting avenues for future research where different automatic metrics could be combined to take advantage of their relative strengths.

\section{Related Work}

Some of the related work, particularly the relevant datasets, human evaluation criteria,  and automatic metrics were already discussed earlier and hence not covered again here.
We refer the readers to two recent surveys \cite{DBLP:journals/corr/abs-2008-12009_NLG_evaluation_metrics_survey,DBLP:journals/corr/abs-2006-14799_other_nlg_eval_survey} for a detailed overview of automatic evaluation metrics as well as related work on criticising the use of automatic evaluation metrics. We mention a few such important works here. BLEU is one of the most widely analysed metric with several studies showing that it does not correlate well with human judgements for machine translation \cite{callison-burch-etal-2006-evaluating_bleu_in_mt}. 
This issue of poor correlations of metrics with human judgements has been reported on not just BLEU, but also on various other metrics, across several NLG tasks including Question Generation \cite{DBLP:conf/emnlp/NemaK18_qbleu}, Data-to-Text generation \cite{DBLP:conf/acl/DhingraFPCDC19_parent}, Dialogue generation \cite{DBLP:conf/emnlp/LiuLSNCP16HowNot}, and Summarisation \cite{kryscinski-etal-2019-neural_summ_critic}. Apart from poor correlations, 
\citet{kryscinski-etal-2019-neural_summ_critic} criticize the automatic metrics for abstractive summarization since they don't check for factual inconsistencies in the summaries. Similarly \citet{DBLP:conf/emnlp/WisemanSR17} discuss the lack of a reliable measurement of faithfulness in the context of Data-to-Text Generation.  %
In case of dialogue, several n-gram-based and embedding-based metrics have been shown to fall short in capturing the diversity of the valid responses \cite{DBLP:conf/emnlp/LiuLSNCP16HowNot,DBLP:journals/tacl/SaiMAK20_DDpp}. The alternative of trained metrics, such as ADEM have been shown to be susceptible to adversarial attacks \cite{DBLP:conf/aaai/SaiGKS19Reeval}.

Similar to the main message of our work, some recent works have also called for a more robust evaluation of automatic evaluation metrics \cite{DBLP:journals/corr/abs-1804-11225,DBLP:conf/acl/MathurBC20_tangled}. \citet{DBLP:conf/emnlp/EthayarajhJ20_utility} also critically examine the current approaches towards NLP leaderboards and point towards having multiple metrics along different dimensions such as fairness, efficiency, robustness, etc.

\section{Conclusion}
We conduct a large-scale study involving 6 tasks, 25 automatic evaluation metrics and 18 human evaluation criteria and observe that (i) different criteria such as fluency, coverage, etc are often not correlated and (ii) %
existing metrics have a %
low correlation with most criteria across different tasks. Based on these observations, we suggest an alternative framework for evaluating evaluation metrics which goes beyond computing correlations with the human scores for overall quality. More specifically, we propose perturbation templates which allow a more fine-grained evaluation of such metrics and help in understanding their strengths and more importantly their limitations. We hope that future work on designing evaluation metrics will use our perturbation checklist for evaluating the effectiveness of the proposed metric in assessing different relevant criteria.

\section*{Acknowledgements}
We thank the annotators for helping us evaluate and annotate the perturbations. We thank the anonymous EMNLP-21 reviewers whose comments and feedback helped enhance the paper. We thank the Google India Ph.D. Fellowship Program and the Prime Minister’s Fellowship Scheme for Doctoral Research for supporting Ananya Sai and Samsung IITM-Pravartak Undergraduate Fellowship for supporting Dev Yashpal Sheth. Finally, we thank the Robert Bosch Centre for Data Science and AI for supporting this work.

\appendix

\section{Criteria correlations}
\label{sec:appendix_cri_cri_corrs}

\begin{figure}[!th]
\centering
\begin{subfigure}{.4\textwidth}
  \centering
    \includegraphics[width=0.85\textwidth]{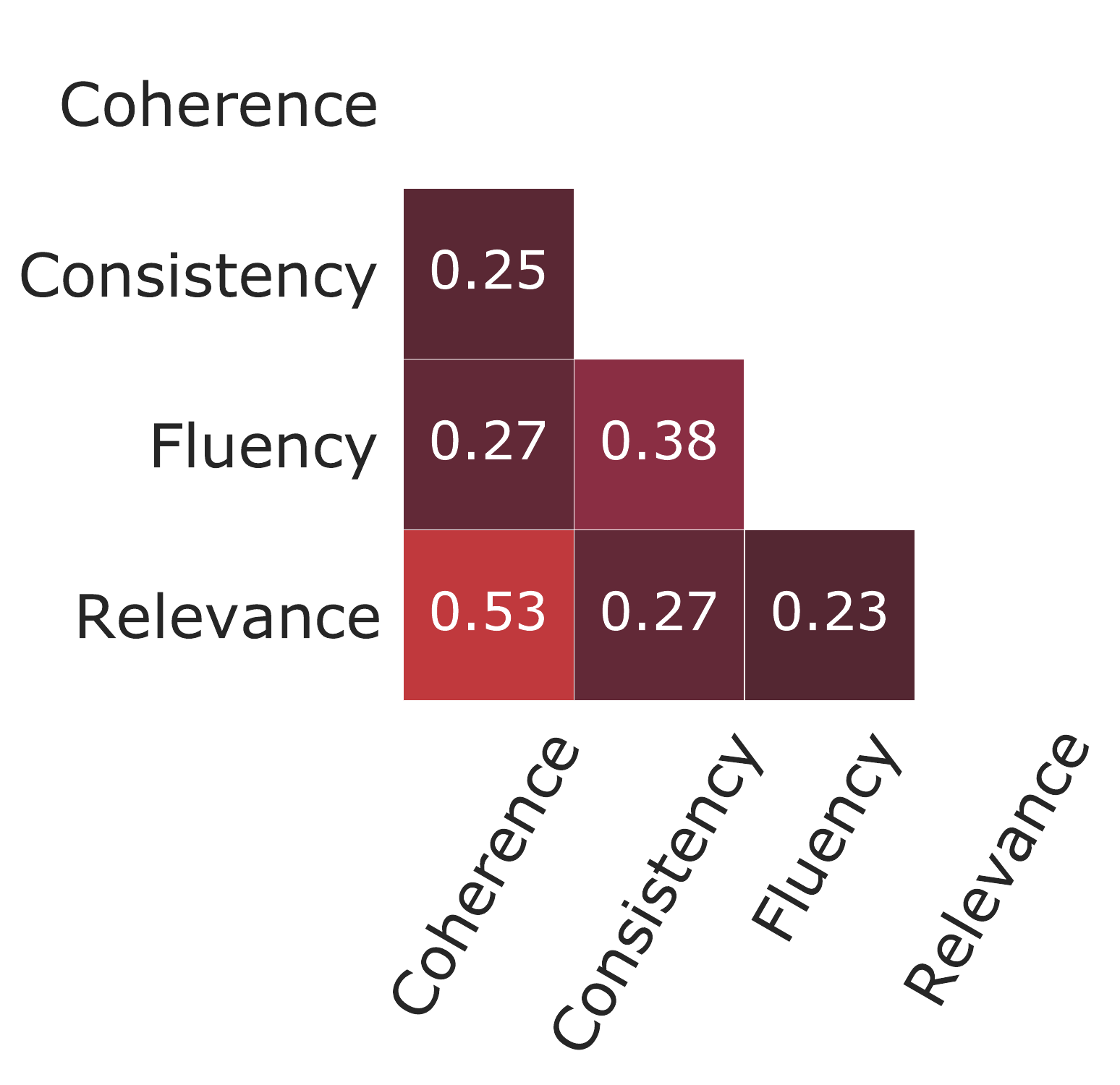}
  \caption{Expert annotations}
  \label{fig:sub_as1}
 \end{subfigure}
\begin{subfigure}{.4\textwidth}
  ~~~ \includegraphics[width=0.85\textwidth]{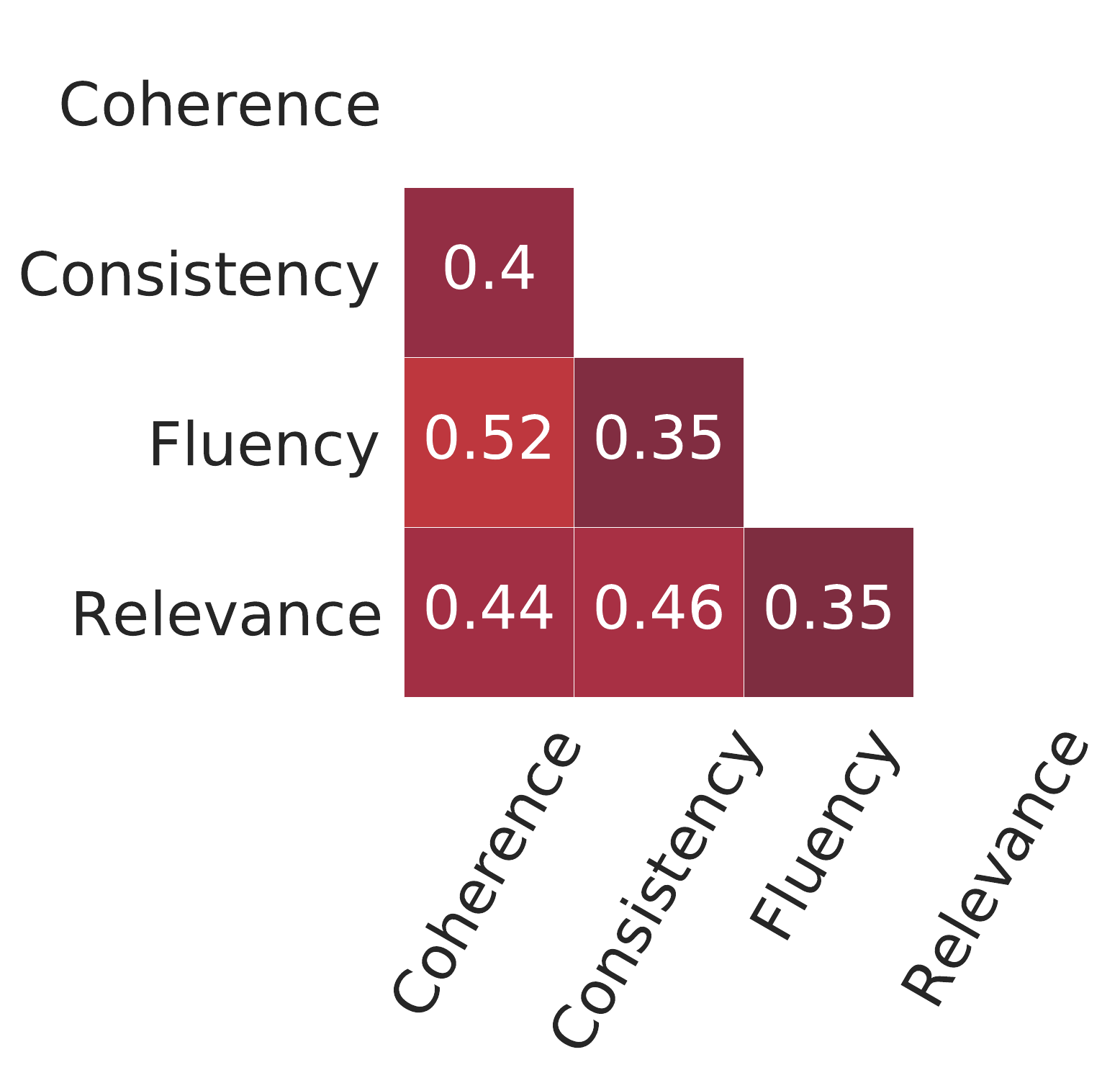}
  \caption{Crowdsourced annotations}
  \label{fig:sub_as2}
\end{subfigure}
\caption{Expert v/s Crowsourced annotations}
\label{fig:suppl_expert_vs_crowd_correlations}
\end{figure}

\begin{figure*}[!th]
\centering
\begin{subfigure}{.3\textwidth}
  \centering
    \includegraphics[width=0.85\textwidth]{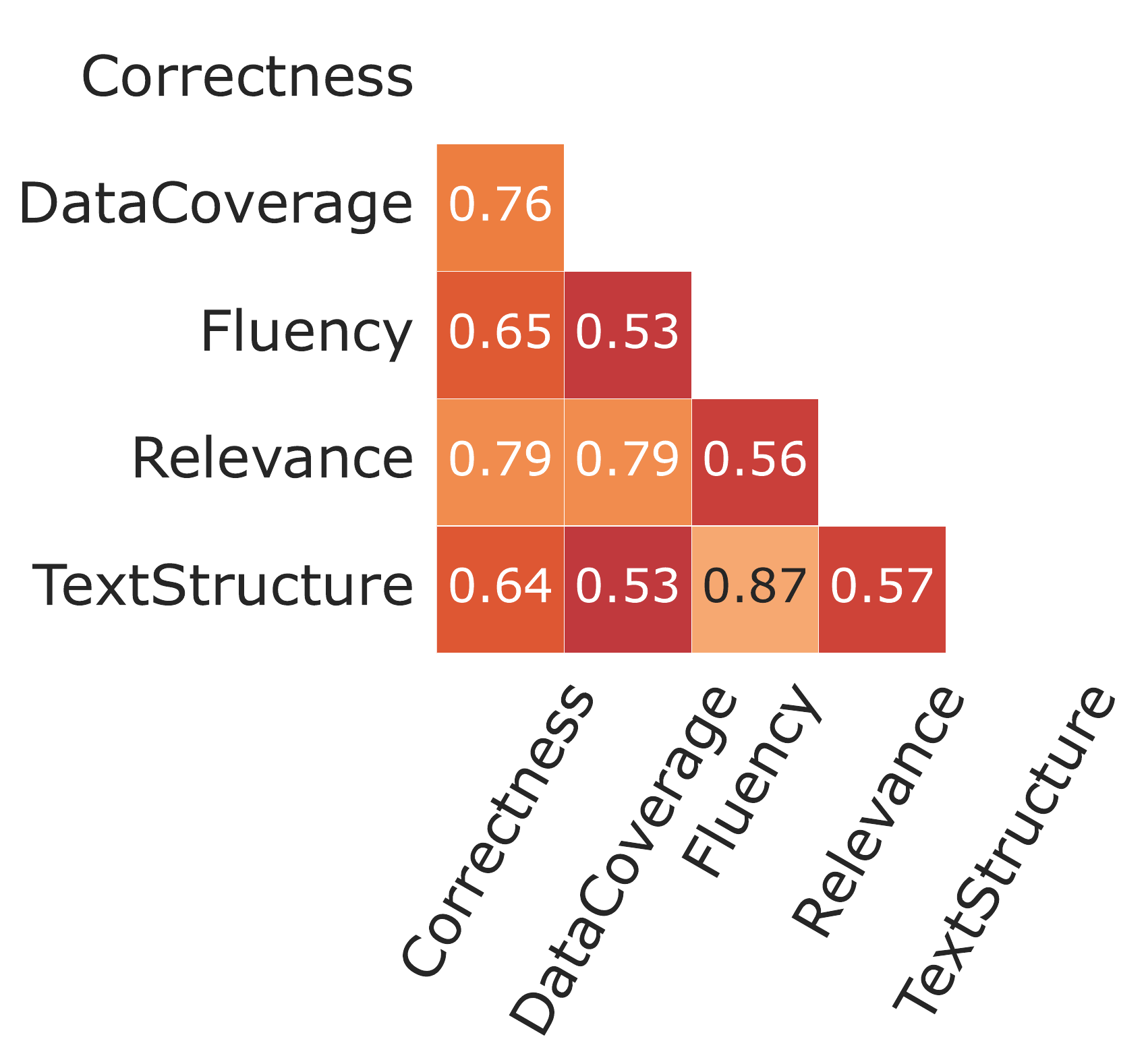}
  \caption{Data to Text Generation criteria}
  \label{fig:sub2_qg}
  ~\\
  ~\\
  ~\\
  \includegraphics[width=0.4\textwidth]{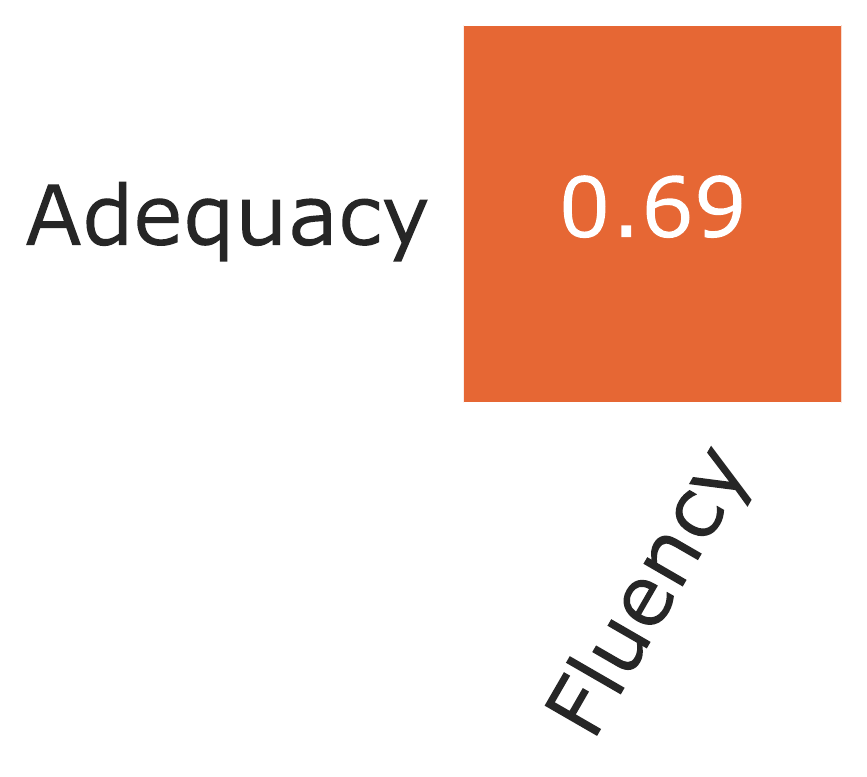}
  \caption{Machine Translation criteria}
  \label{fig:sub2_mt}
 
\end{subfigure}
\begin{subfigure}{0.35\textwidth}
  \includegraphics[width=0.65\textwidth]{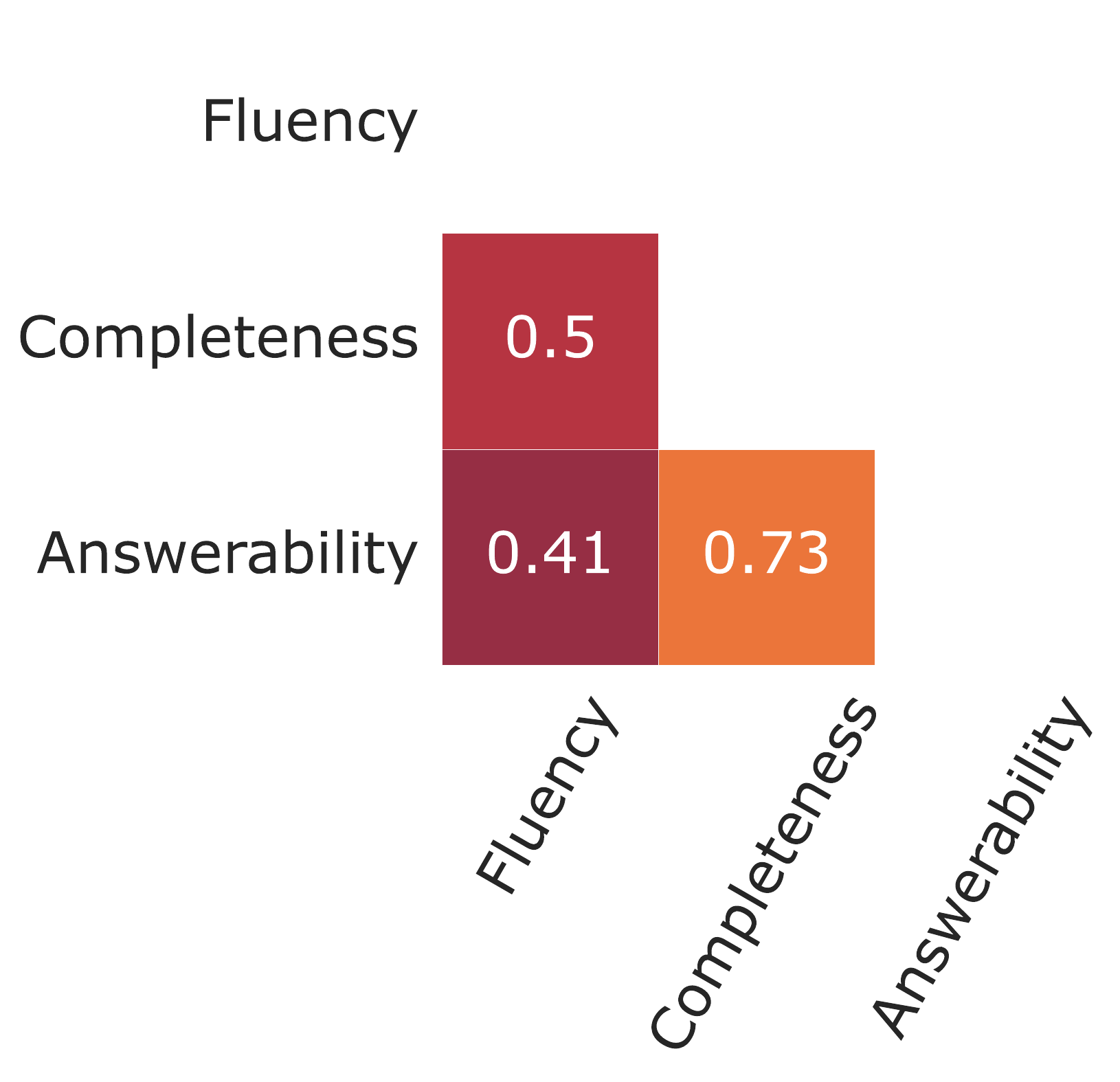}
  \caption{Question Generation criteria}
  \label{fig:sub2_dg}
  ~\\
  \includegraphics[width=1\textwidth]{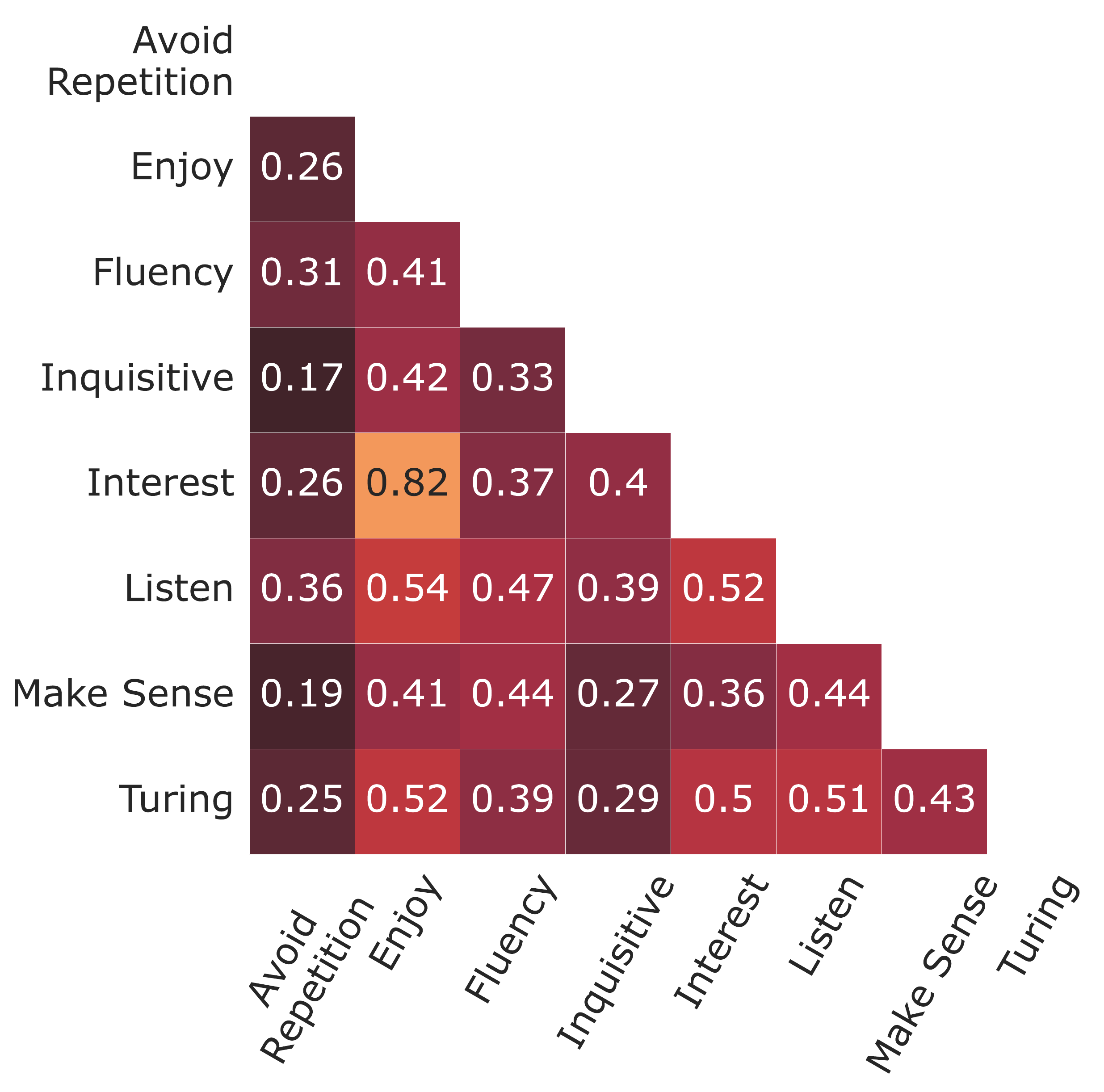}
  \caption{Dialogue Generation criteria}
  \label{fig:sub2_d2t}
\end{subfigure}%
\begin{subfigure}{.3\textwidth}
    \includegraphics[width=0.85\textwidth]{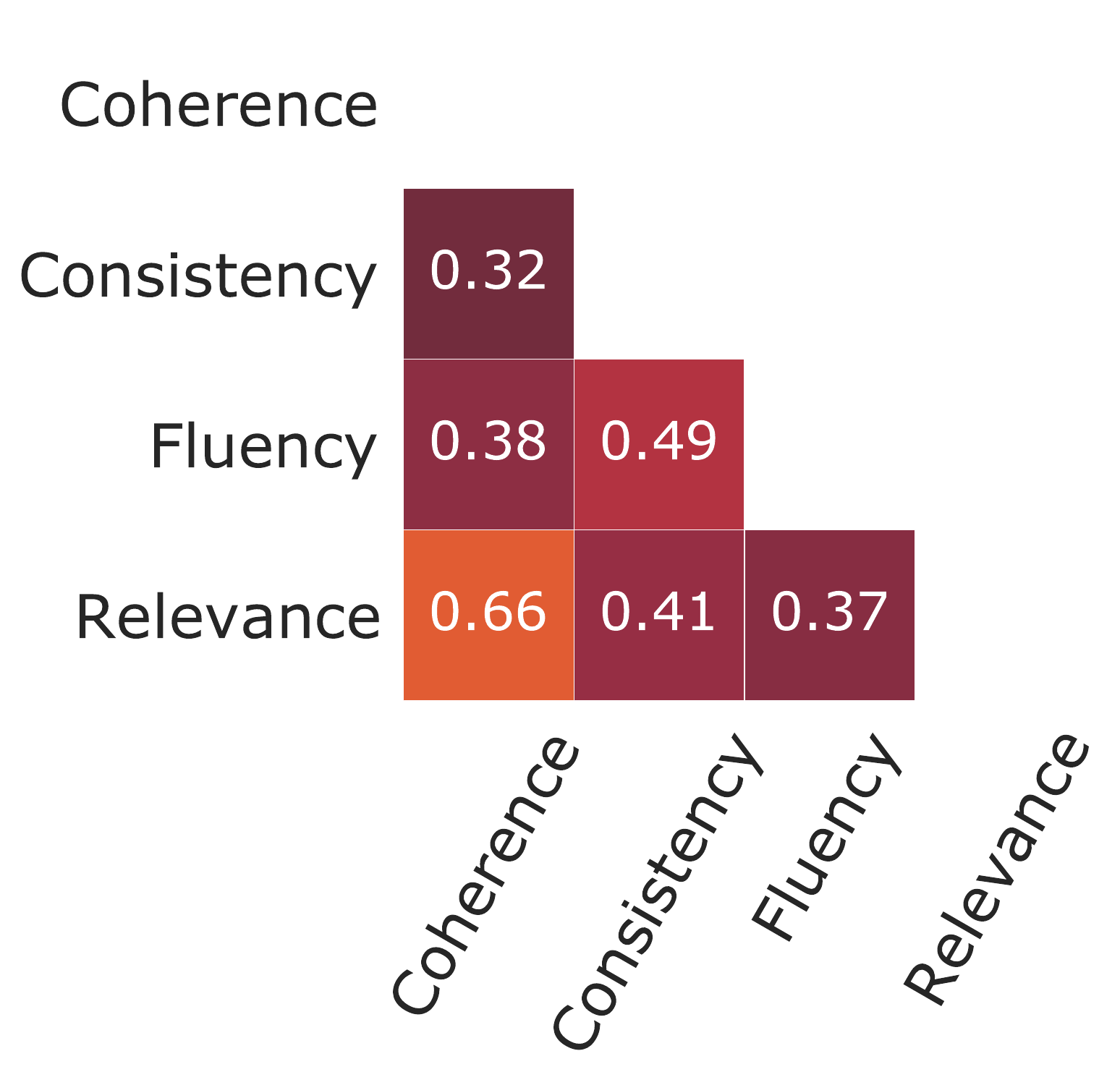}
  \caption{Abstractive Summarisation criteria}
  \label{fig:sub2_as}
  ~\\
  ~\\
  \centering
  \includegraphics[width=0.6\textwidth]{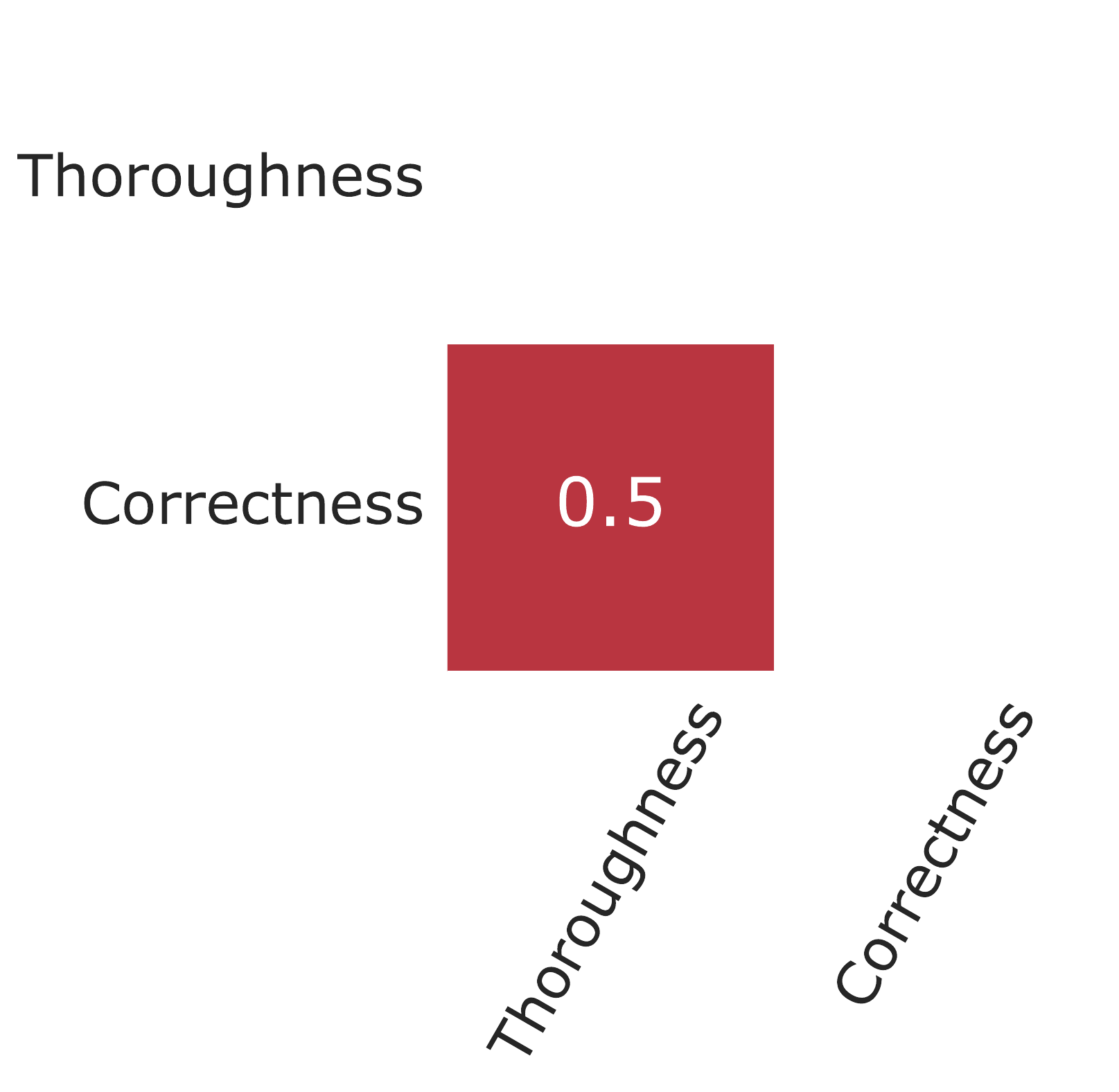}
  ~\\
  \caption{Image Captioning criteria}
  \label{fig:sub2_ic}
  
\end{subfigure}
\caption{Correlations between criteria for 6 different tasks (MT, DG, AS, QG, D2T and IC)}
\label{fig:suppl_all_criteria_correlations}
\end{figure*}

The pearson correlations among the criteria are presented in Figure \ref{fig:suppl_all_criteria_correlations}. Most of the correlation ranges are similar for pearson correlation and kendall tau correlation, except for D2T task. We refer the studies on such correlations \cite{DBLP:conf/acl/MathurBC20_tangled}, discussing various points such as the influence of outliers and noisy points on the correlations. Additionally, we observe that the expertise of the annotators also influences the criteria-criteria correlations. In particular, we were able to study this in case of AS using the data released by \citet{DBLP:journals/corr/abs-2007-12626_summeval} containing both expert and crowdsourced annotations. From Figure \ref{fig:suppl_expert_vs_crowd_correlations}, we observe that the scores by expert annotators have far lesser correlations amidst various criteria than the crowdsourced annotations.

\section{Detailed examples to illustrate the implementation of the perturbation templates}
\label{sec:appendix_perturbation_details}
Our perturbation templates mainly draw from the official github repository\footnote{\href{https://github.com/marcotcr/checklist}{https://github.com/marcotcr/checklist}} of the checklist paper and are also publicly available\footnote{\href{https://github.com/iitmnlp/EvalEval}{https://github.com/iitmnlp/EvalEval}}. The implementation involves preprocessing with the help of tokenization, POS tagging, NER recognition, etc. Synonyms, antonyms, etc., are obtained with the help of WordNet framework. Additionally, the masked language model of RoBERTa is used to mask and predict replacements for the targeted words. For example, the application of the template for `dropping stop words' involves the tokenization of the sentence using the NLTK word tokenizer as the first step. The list of tokens is compared with the set of stopwords provided by NLTK to filter out the stop words from the list of tokens. The modified sentence is then reconstructed using the string join function by iterating over the tokens in the modified list. Similarly, for the template of `changing the attributes' in case of image captioning, the sentence is first tokenized, then the adjectives are identified using part-of-speech tagging (again a functionality provided by NLTK). The list of `related words' (i.e., hyponyms of hypernyms or `sibling words') are obtained using WordNet framework. Unless the list returns empty, one of the entries in the list is used to replace the original adjective. In order to `change question to an assertive statement', the question words (such as who, what, why, when, etc) are replaced with a `mask' token and the `?' character at the end is replaced with `.' using string replace function. This modified sentence is then fed to RoBERTa model which generates different predictions to be used in place of the `{mask}' token. One of the suggested words is used to form the modified assertive sentence. In case of perturbations involving dropping words, we additionally decide if we're dropping stop words, adjectives, question words, etc in the particular perturbation and estimate the extent of effect it'll have on each criteria. %
The perturbations of adding text, appends random words / phrases / sentences to a given text to account for not just the cases where there is missing information, but also cases where there is spurious wrong information, even if it accompanies / follows the correct version. 
The complete implementations / details of our perturbation templates are hosted publicly\footnote{\href{https://github.com/iitmnlp/EvalEval}{https://github.com/iitmnlp/EvalEval}}. %

Note that some of the perturbations cannot be applied to every sentence in the dataset. For example, the template of ``changing names'' cannot be applied if there are no named entities in a particular sentence. We hence shortlist only the successfully modified samples from the dataset for analysing the metrics' performance on each perturbation. 

\section{Perturbation Templates for various criteria}
\label{sec:appendix_full_templates}
Table \ref{table:examples_suppl} contains the comprehensive list of perturbation templates used in our work. 
\begin{table*}[!t]
\centering %
\resizebox{1\textwidth}{!}{
\rowcolors{1}{}{lightgray!50}
\begin{tabular}{l L{2cm} L{4cm} L{7.5cm} L{7.5cm}} %
\hline\hline %
\textbf{Task} & \textbf{Criteria} & \textbf{Perturbation} & \textbf{Unmodified sentence} & \textbf{Perturbed sentence} \\ %
\hline \hline%

 \cellcolor{white}   &\cellcolor{white}  &Misplaced/missing punctuation&Could you let me know if I can meet him now or later\colorbox{color}{?}& Could you let me know\colorbox{lightred}{,} if I can meet him now or later\colorbox{lightred}{.}\\
 \cellcolor{white}   &\cellcolor{white}& Jumbling word order & We \colorbox{color}{play} badminton every evening. & We badminton every evening play. \\
    \cellcolor{white}   &\cellcolor{white} &  Subject-verb disagreement & He \colorbox{color}{doesn't} know how to bake. & He \colorbox{lightred} {don't} know how to bake.\\
    \cellcolor{white}   &\cellcolor{white} &Dropping words (such as prepositions/articles, etc)& \colorbox{color}{The} bank \colorbox{color}{is} willing to approve the loan. & Bank willing to approve the loan. \\
    \multirow{-8}{1.4em}{\cellcolor{white}All tasks} &  \multirow{-8}{*}{\cellcolor{white}Fluency}  &Spelling errors&Make the most of every opportunity presented to you.&Make the most of \colorbox{lightred}{evry} opportunity presented to you.\\
    \hline

    \cellcolor{white}   &\cellcolor{white}    &Dropping out words or phrases &I \colorbox{color}{was being} followed. & I followed. \\ %
    \cellcolor{white}   &\cellcolor{white} &Adding extra wrong information& This book is so inspiring. & This book is so inspiring,\colorbox{lightred} {I forgot}. \\ 
    \cellcolor{white}   &\cellcolor{white} &Negation / antonyms&It \colorbox{color}{will rain} on Monday.&It will \colorbox{lightred}{not} rain on Monday.\\
    \multirow{-7}{1.4em}{\cellcolor{white}MT} & \multirow{-7}{*}{\cellcolor{white}Adequacy}  &Repeat phrases &My relatives are in town.&My relatives are in town, \colorbox{lightred}{my relatives}.\\
    \hline

\cellcolor{white} & \cellcolor{white}  
    &Dropping words& Here is the \colorbox{color}{no parking} sign. & Here is the sign.\\ %
    \cellcolor{white}   &\cellcolor{white} &Negation and antonyms& This book is so \colorbox{color}{inspiring}. & This book is so \colorbox{lightred}{uninspiring}. \\
    \cellcolor{white} &\multirow{-4}{4em}{\cellcolor{white}Informa-tiveness} &Use hyponyms to create misinformation& The girl my \colorbox{color}{brother} Andy met through MySpace turned out to be completely made up . & The girl my \colorbox{lightred}{friend} Andy met through MySpace turned out to be completely made up. \\ 
    \hhline{~----}\hhline{~}
    \cellcolor{white} & \cellcolor{white}Flow / coherence & Reorder sentences & The pandemic was spreading uncontrollably. Vaccines are being developed and tested rapidly.& Vaccines are being developed and tested rapidly. \colorbox{lightred}{The pandemic was spreading uncontrollably.}\\
    \hhline{~----}\hhline{~}
    \cellcolor{white} &\cellcolor{white} Non-Redundancy & Repeat sentences& My relatives are in town. &My relatives are in town. \colorbox{lightred}{My relatives in town.}\\
    \hhline{~----}\hhline{~}
\multirow{-9}{*}{\cellcolor{white}AS}  &\cellcolor{white}Referential clarity& Replace nouns by pronouns& \colorbox{color}{The pandemic} was spreading uncontrollably. \colorbox{color}{Vaccines} are being developed rapidly.&\colorbox{lightred}{It} was spreading uncontrollably. \colorbox{lightred}{They} are being developed rapidly.\\
    \hline

\cellcolor{white} & \cellcolor{white} 
    &Change question word& \colorbox{color}{When} was he born? &\colorbox{lightred}{What} was he born? \\ %
    \cellcolor{white} & \cellcolor{white}  &Drop question word& \colorbox{color}{When} was he born ? & Was he born?\\
    \cellcolor{white} & \cellcolor{white}  &Change question to assertive statement & Who is the director of Titanic? & \colorbox{lightred}{The director of Titanic is James Cameron.}\\
    \cellcolor{white} & \multirow{-5}{4em}{\cellcolor{white}Answera-bility}   &Drop NLTK stop words&Who \colorbox{color}{was not a} world leader&Who world leader?\\
    \hhline{~----}\hhline{~}
    \cellcolor{white} & \cellcolor{white}  &Mask a few words and predict&How could Tesla run \colorbox{color}{the experiments}?&How could Tesla run to \colorbox{lightred}{the beach}?\\
 \multirow{-9}{1.4em}{\cellcolor{white}QG}    & \multirow{-4}{4em}{\cellcolor{white}Relevance} &Perturb named entities, nouns, verbs& Why did \colorbox{color}{Mary} go to the \colorbox{color}{school}? & Why did \colorbox{lightred}{Raj} go to the \colorbox{lightred}{market}? \\
 \hline

\cellcolor{white} & \cellcolor{white}
    &{\small Negate a previous statement by same speaker}& Bot: I enjoy having your daughter in my class. User: I'm glad to hear that. & I \colorbox{lightred}{don't} enjoy having your daughter in my class.\\ %
    \cellcolor{white} & \multirow{-3}{4em}{\cellcolor{white}Making sense} &Add extra non-sensible text & Do you know where Dr. XYZ lives? & Yes, \colorbox{lightred}{my father is my grandmother's father} \\
    \hhline{~----}\hhline{~}
    \cellcolor{white} & \cellcolor{white}
    & Repeat previous utterances  & Do you know Dr.XYZ?& \colorbox{lightred}{Do you know where Dr. XYZ lives?}\\
    \cellcolor{white} & \multirow{-3}{4em}{\cellcolor{white}Avoid repetition} & Repeat phrases & I like ice creams& I like ice creams, \colorbox{lightred}{ice creams}\\
    \hhline{~----}\hhline{~}
    \cellcolor{white} & \cellcolor{white}Listening & Replace with "Can you repeat?" & I need to book a taxi & \colorbox{lightred}{I'm sorry, can you repeat?}\\
    \hhline{~----}\hhline{~}
    \multirow{-11}{*}{\cellcolor{white}DG} & \cellcolor{white}Relevance & Map random responses & I am \colorbox{color}{new to coding}. & I am \colorbox{lightred}{scared of snakes}.\\
    \hline
    
\cellcolor{white} & \cellcolor{white} & Change the order of objects & The \colorbox{color}{man} is standing in front of a \colorbox{color}{tree} & The tree is standing in front of man \\
    \cellcolor{white} & \cellcolor{white}
    & Change gender  &  Two \colorbox{color}{girls} are playing with a doll & Two \colorbox{lightred}{boys} are playing with a doll\\
    \cellcolor{white} & \multirow{-3}{3em}{\cellcolor{white}Correctness} & Change attributes  &  A \colorbox{color}{small} boy playing with a \colorbox{color}{red} ball & A \colorbox{lightred}{tall} boy playing with a \colorbox{lightred}{green} ball\\
    \hhline{~----}\hhline{~}
    \cellcolor{white} & \cellcolor{white}
    & Drop objects/noun  & A small \colorbox{color}{boy} playing with a red \colorbox{color}{ball} &  A small playing with a red \\
    \multirow{-6}{*}{\cellcolor{white}IC} & \multirow{-2}{4em}{\cellcolor{white}Thoroughness} & Repeat (append) object & A lady riding a horse. & A lady riding a horse \colorbox{lightred}{and a lady}.\\
    \hline

    \cellcolor{white} & \cellcolor{white} & Use hyponyms & Beethoven was a German \colorbox{color}{musician} & Beethoven was a German \colorbox{lightred}{architect} \\
    \cellcolor{white} & \cellcolor{white}
    & Change numbers  &  The cricketer was born in \colorbox{color}{1990}. & The cricketer was born in \colorbox{lightred}{1950}.\\
    \cellcolor{white} & \multirow{-3}{4em}{\cellcolor{white}Correctness} & Negation / antonyms &  The author of Harry Potter is J.K Rowling & The author of Harry Potter is \colorbox{lightred}{not} J.K Rowling.\\
    \hhline{~----}\hhline{~}
    \cellcolor{white} & \cellcolor{white}
    & Drop phrases  & A small boy playing with a \colorbox{color}{red ball} &  A small boy playing with a \\
    \cellcolor{white} & \multirow{-3}{4em}{\cellcolor{white}Data Coverage} & Repeat phrases & Beethoven was a German musician & Beethoven was a German musician \colorbox{lightred}{and German musician}.\\
    \hhline{~----}\hhline{~}
    \cellcolor{white} & \cellcolor{white}
    & Random text& Beethoven was a German musician & \colorbox{lightred}{The cricketer was born in 1990.}\\ %
    \multirow{-10}{*}{\cellcolor{white}D2T}& \cellcolor{white}Relevance & Perturb names & \colorbox{color}{Phillips} was a child prodigy. & \colorbox{lightred}{James} was a child prodigy.\\
    \hline

\cellcolor{white} & \cellcolor{white}
    &Replace with synonyms&The mangoes are \colorbox{color}{delicious}. & The mangoes are \colorbox{lightred}{tasty}.\\ %
    \cellcolor{white} & \cellcolor{white}&Contractions&We \colorbox{color}{are} going to embark on an adventure. & \colorbox{lightred}{We're} going to embark on an adventure.\\ 
    \cellcolor{white} & \cellcolor{white}&Expansions&There \colorbox{color}{weren't} any clear winners of the contest&There \colorbox{lightred}{were not} any clear winners of the contest.\\
    \multirow{-5}{1.4em}{\cellcolor{white}All tasks}   & \multirow{-5}{*}{\cellcolor{white}Invariance} &Numerals to words&Aron Ralston who was trapped for \colorbox{color}{127} hours.&Aron Ralston who was trapped for \colorbox{lightred}{one hundred twenty seven} hours.\\
    \cline{1-4}
\hline
\end{tabular}}
\caption{Perturbation templates targeting various criteria with examples. The blue highlights indicate the portions of the original sentence affected by the perturbation template. The red highlights indicate the changes in the modified sentence.}
\label{table:examples_suppl} %
\end{table*}

\section{Correlations of Metrics with various Criteria (complete plots)}
\label{sec:appendix_full_plots}
\begin{figure*}[!ht]
\begin{subfigure}{1\textwidth}
    \centering
    \includegraphics[width=1\textwidth]{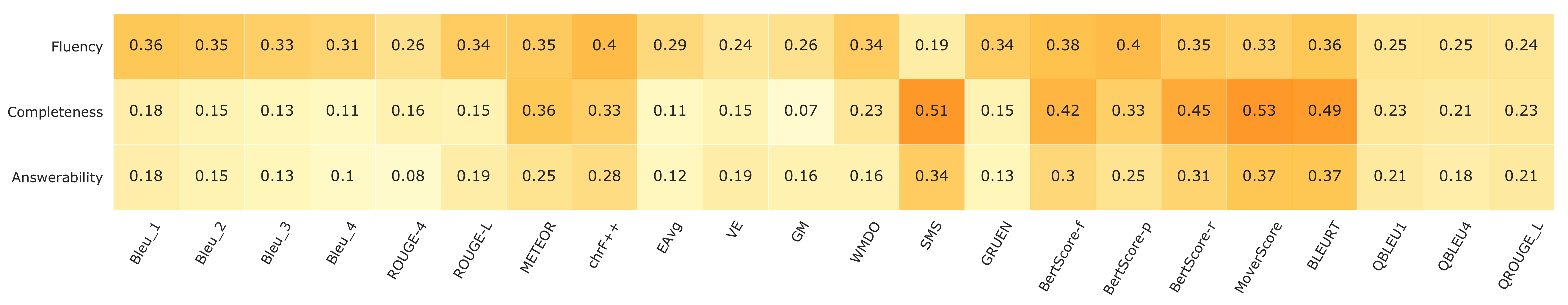}
    \vspace{-11mm}
    \caption{Question Generation (QG) }
    \label{fig:sqg_met_corrs}
    \centering
    \includegraphics[width=1\textwidth]{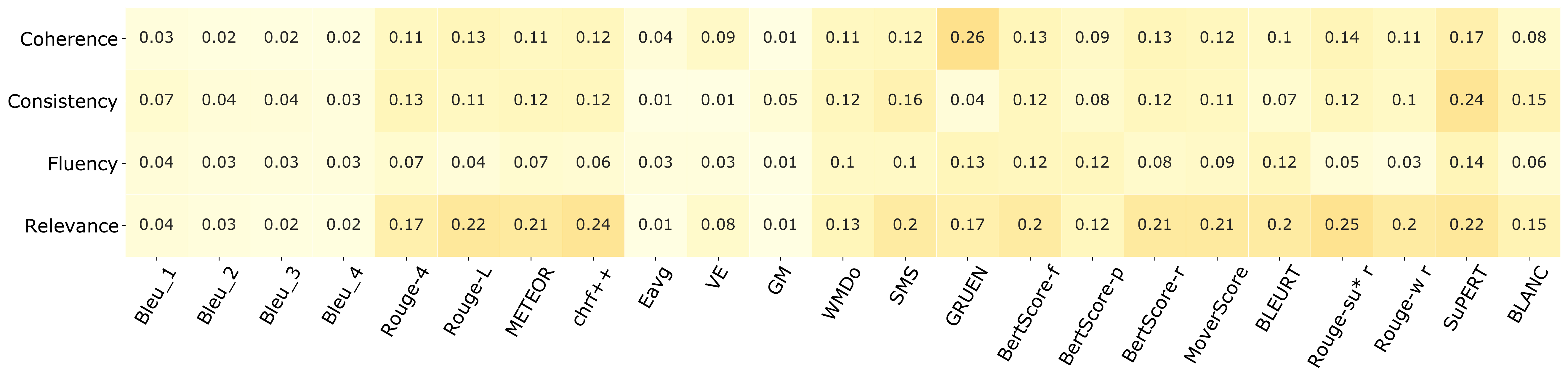}
    \vspace{-9.5mm}
    \caption{Abstractive Summarization (AS) }
    \label{fig:sas_met_corrs}
    \centering
    \includegraphics[width=1\textwidth]{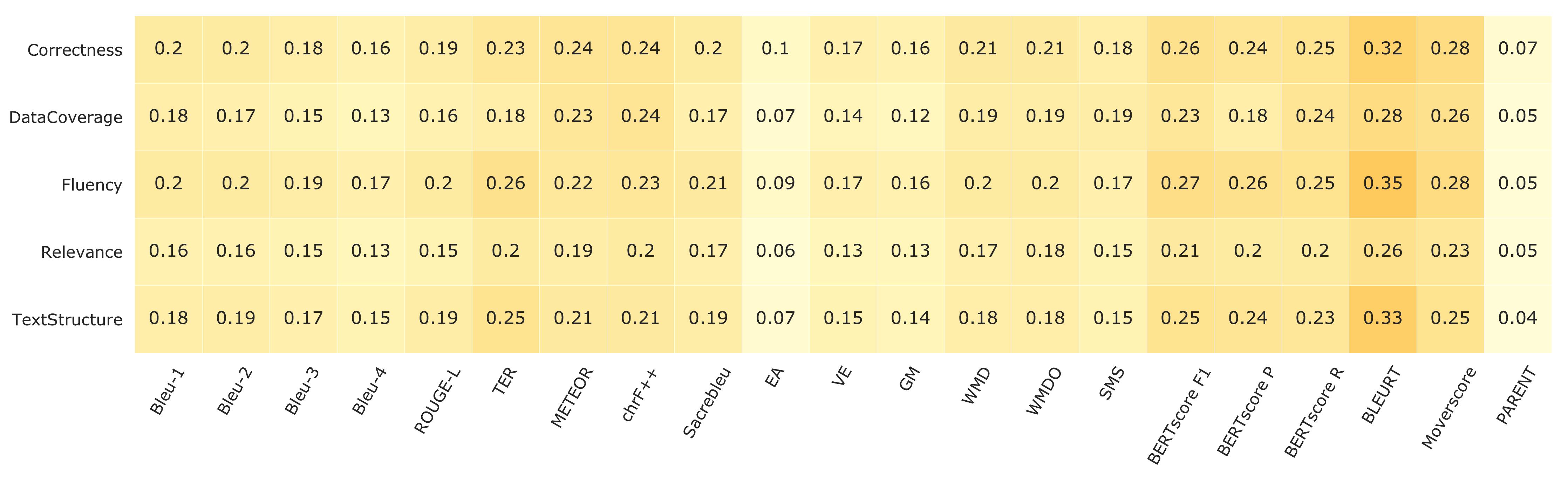}
    \vspace{-11mm}
    \caption{Data to Text Generation (D2T) }
    \label{fig:sdt_met_corrs}
    \centering
    \includegraphics[width=1\textwidth]{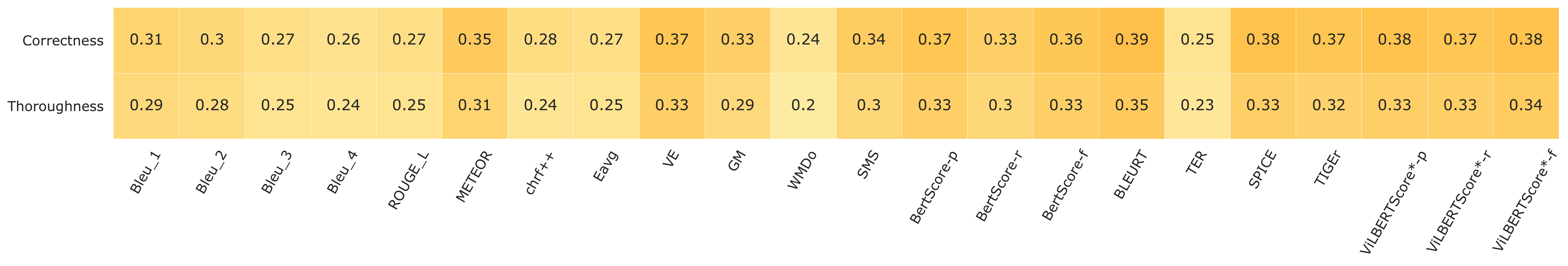}
    \vspace{-11mm}
    \caption{Image Captioning }
    \label{fig:sic_met_corrs}
    \includegraphics[width=0.5\textwidth]{imgs/final_plots/dg_kendall.pdf}
    \vspace{-3mm}
    \caption{Dialogue Generation (DG)}
    \label{fig:sdg_met_corrs}
\end{subfigure}
\caption{Correlations of metrics with different criteria}
\label{fig:supple_all_metric_criteria_correlations}
\end{figure*}

Figure \ref{fig:supple_all_metric_criteria_correlations} is a more comprehensive version of Figure \ref{fig:all_metric_criteria_correlations}. It shows the correlations of the complete set of metrics considered in this study with various criteria across different tasks.

\end{document}